\documentclass[runningheads]{llncs}
\usepackage{graphicx}

\usepackage{tikz}
\usepackage{comment}
\usepackage{amsmath,amssymb} 
\usepackage{color}

\usepackage{multicol}
\usepackage{multirow}
\usepackage{booktabs}
\usepackage{makecell}
\usepackage{adjustbox}

\usepackage[T1]{fontenc}

\usepackage[symbol]{footmisc}

\usepackage[accsupp]{axessibility}  

\definecolor{ben}{rgb}{0.9,0.,0.5}

\definecolor{pat}{rgb}{0.6,0.2,0.1}

\definecolor{GY}{rgb}{0.4,0.1,0.9}

\newlength\savewidth\newcommand\shline{\noalign{\global\savewidth\arrayrulewidth
  \global\arrayrulewidth 1pt}\hline\noalign{\global\arrayrulewidth\savewidth}}

\begin{document}
\pagestyle{headings}
\mainmatter
\def\ECCVSubNumber{2796}  

\title{Is my Depth Ground-Truth Good Enough? HAMMER - Highly Accurate Multi-Modal Dataset for DEnse 3D Scene Regression}


\titlerunning{HAMMER dataset}
%
\author{HyunJun Jung\inst{1} \and
Patrick Ruhkamp\inst{1} \and
Guangyao Zhai\inst{1} \and
Nikolas Brasch\inst{1} \and
Yitong Li\inst{1} \and
Yannick Verdie\inst{2} \and
Jifei Song\inst{2} \and
Yiren Zhou\inst{2} \and
Anil Armagan\inst{2} \and
Slobodan Ilic\inst{1,3} \and
Ales Leonardis\inst{2} \and
Benjamin Busam\inst{1}
\\ \footnotesize{\fontfamily{qcr}\selectfont
hyunjun.jung@tum.de, b.busam@tum.de
}
}

\authorrunning{HJ. Jung et al.}
%
\institute{Technical University of Munich\and Huawei Noah's Ark Lab, \ \ \ \inst{3} \ Siemens AG}

\maketitle

\begin{abstract}
Depth estimation is a core task in 3D computer vision. Recent methods investigate the task of monocular depth trained with various depth sensor modalities. Every sensor has its advantages and drawbacks caused by the nature of estimates. In the literature, mostly mean average error of the depth is investigated and sensor capabilities are typically not discussed. Especially indoor environments, however, pose challenges for some devices. Textureless regions pose challenges for structure from motion, reflective materials are problematic for active sensing, and distances for translucent material are intricate to measure with existing sensors. This paper proposes HAMMER, a dataset comprising depth estimates from multiple commonly used sensors for indoor depth estimation, namely ToF, stereo, active stereo together with monocular RGB+P data. We construct highly reliable ground truth depth maps with the help of 3D scanners and aligned renderings. A popular depth estimators is trained on this data and typical depth sensors. The estimates are extensively analyze on different scene structures. We notice generalization issues arising from various sensor technologies in household environments with challenging but everyday scene content. HAMMER, which we make publicly available (https://github.com/Junggy/HAMMER-dataset), provides a reliable base to pave the way to targeted depth improvements and sensor fusion approaches.
\end{abstract}

\begin{figure*}[tbp]
 \centering
    \includegraphics[width=\linewidth]{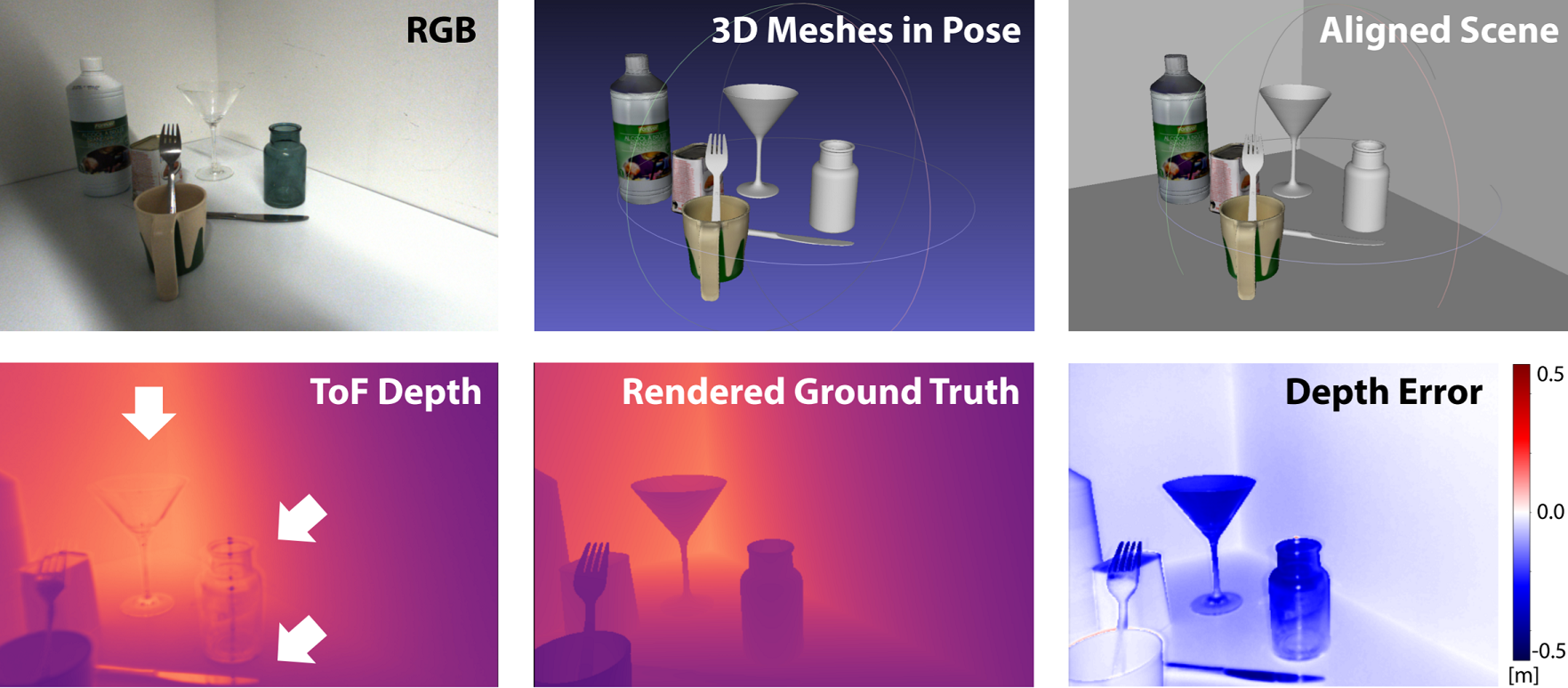}
    \caption{We acquire indoor scenes comprising photometrically challenging everyday objects with multiple modalities. In order to acquire highly accurate and reliable ground truth depth maps, we align detailed pre-recorded 3D meshes with the acquisition (upper center) and the scene (upper right). Acquired depth sensor data (exemplified with ToF in the lower left) faces issues, especially for reflective and translucent objects (lower right), for which we provide highly accurate ground truth depth maps (lower center).}
    \label{fig:teaser}
\end{figure*}

\section{Introduction}
\label{sec:introduction}

Our world is 3D. Distance measurements are essential for machines to understand and interact with our environment spatially.
Autonomous vehicles~\cite{geiger2012we,ruhkamp2021attention} need this information to drive safely, and robot vision requires distance information to manipulate objects~\cite{fang2020graspnet,wang2021demograsp}, and AR realism benefits from spatial understanding~\cite{kopf2021robust,busam2019sterefo}.


A wide variety of sensor modalities, visual distance measurement technologies, and depth prediction pipelines exist.
The computer vision community thereby benefits from a wide diversity of publicly available datasets~\cite{scharstein2002taxonomy,geiger2012we,silberman2012indoor,sturm2012benchmark,xiao2013sun3d,PhoCal,CroMo}, which allow for the objective evaluation of depth estimation pipelines.

Depending on the setup, different sensors are chosen to provide ground truth depth maps, all of which have their respective advantages and drawbacks determined by distance reasoning.
Pipelines are usually trained on the data without questioning the nature of the depth sensor and do not reflect areas of high or low confidence.

Popular \textbf{passive sensors} setups include multi-view stereo cameras where the known or calibrated spatial relationship between them is used for depth reasoning~\cite{scharstein2002taxonomy}. 
Corresponding image parts or patches are photometrically or structurally associated, and geometry allows to triangulate points within an overlapping field of view.
Such photometric cues are not reliable in low-textured areas and with little ambient light where \textbf{active sensing} can be beneficial~\cite{silberman2012indoor,sturm2012benchmark}.

Active stereo can be used to artificially create texture cues in low-textured areas and photon-pulses with a given sampling rate are used in Time-of-Flight (ToF) setups either directly from the pinhole (dToF) or another perspective indirectly (iToF)~\cite{Guo_2018_ECCV}.
With the speed of light, one can measure the distance of objects from the return time of the light pulse, but unwanted multi-reflection artifacts also arise.
Reflective and translucent materials are measured at incorrect far distances, and multiple light bounces distort measurements in corners and edges. 
While ToF signals can still be aggregated for dense depth maps, a similar setup is used with LiDAR sensors which sparsely measure the distance using coordinated rays that bounce from objects in the surrounding.
The latter provides ground truth, for instance, for the popular outdoor driving benchmark KITTI~\cite{geiger2012we}.
While LiDAR sensing can be costly, radar~\cite{gasperini2021r4dyn}  provides an even sparser but more affordable alternative.


Multiple modalities can also be fused to enhance distance estimates.
A common issue, however, is the inherent problem of warping onto a common reference frame which requires the information about depth itself~\cite{lopez2020project,jung2021wild}.
While multi-modal setups have been used to enhance further monocular depth estimation using self-supervision from stereo and temporal cues~\cite{monodepth2,CroMo}, its ground truth analysis is mainly limited to average errors and restricted by the individual ground truth sensor used.

An unconstrained analysis of depth in terms of RMSE compared against a ground truth sensor only shows part of the picture as different sensing modalities may suffer from drawbacks.
Where are the drawbacks of current depth-sensing modalities - and how does this impact pipelines trained with this (potentially partly erroneous) data? 
Can self- or semi-supervision overcome some of the limitations posed currently?

To investigate these questions, we build a unique camera rig comprising a set of the most popular indoor depth sensors and acquire synchronized captures with highly accurate ground truth data (see Fig.~\ref{fig:teaser}) using 3D scanners.
We then train typical RGB-based depth estimation pipelines using various ground truth signals from different modalities and evaluate their performance separately on areas of different photometric complexity and with varying structural and material properties.
To this end, our main contributions can be summarized as follows:

\begin{enumerate}
    \item We acquire an \textbf{indoor dataset} comprising an unprecedented combination of \textbf{multi-modal sensors}, namely iToF, dToF, monocular RGB+P, monochrome stereo, and active light stereo. The dataset contains highly accurate ground truth data of 13 indoor scenes with a total of 13k frames. We make the dataset publicly available to open the path to more targeted depth improvements and novel sensor fusion concepts.
    \item We perform a thorough \textbf{monocular depth estimation analysis} to compare the influence of different sensing modalities as supervision signals. For the first time, we separately investigate the performance on textured and untextured areas as well as \textbf{photometrically challenging} reflective, translucent and transparent \textbf{scenes}.
\end{enumerate}
\section{Related Work}
\label{sec:relatedwork}
\subsection{Depth from X}
A variety of sensor modalities have been used to obtain depth maps.
Typical datasets comprise one ground truth sensor used for all acquisitions, which is assumed to give accurate enough data to validate the models:

\smallskip
\noindent
\textit{Stereo Vision.}
In the stereo literature, early approaches~\cite{scharstein2002taxonomy} restricted the scene to piecewise planar objects for which ground truth data was provided to compare disparity estimation methods.
With the requirement of sub-pixel precise annotations, later approaches~\cite{scharstein2014high} leveraged structured light with multi-frame acquisitions as highly reliable ground truth for a set of static scenes.

\smallskip
\noindent
\textit{Structured Light sensing.}
The availability of structured light sensors made it possible to acquire real indoor environments~\cite{silberman2012indoor} where depth data at missing pixels could be inpainted. Structure from motion (SfM) was used to generate the depth maps of Sun3D~\cite{xiao2013sun3d} where a moving camera acquired the scenes.
To further enable temporal data in particular regarding evaluations of SLAM pipelines, mocked-up indoor scenes were used in the pioneering dataset of Sturm et al.~\cite{sturm2012benchmark}.
While advancing the field, its depth maps were limited to the IR-pattern approach used in these RGBD sensors.


\smallskip
\noindent
\textit{ToF Dataset.}
Further advances in active depth sensing emphasize ToF more, where recent methods focus on curing so evoked depth maps from multipath interference and motion artifacts.
Initial investigations focused on simulated data~\cite{Guo_2018_ECCV} and controlled environments with little ambient noise~\cite{son2016learning}.
The broader availability of ToF sensors in commercial products such as the Kinect~v2 and the Kinect Azure as well as modern smartphones (e.g. iToF of Huawei P30 Pro, dToF in Apple iPhone 12) brought this line of research closer to real applications.
More recent studies applied these methods to realistic scenes with a high level of sparsity and shot noise~\cite{jung2021wild} paving the way to mobile ToF sensing.
For this reason, we include dToF and iToF modalities in all our experiments.

\smallskip
\noindent
\textit{Polarimetric Cues.}
Other properties of light are used to indirectly retrieve scene surface properties in the form of normals for which the amount of linearly polarized light and its polarization direction provide some cues, especially for highly reflective and transparent objects~\cite{kalra2020deep,gao2021polarimetric}.
While initial investigations for shape from polarization (SfP) mainly analysed controlled setups~\cite{garcia2015surface,atkinson2006recovery,yu2017shape,smith2018height}, recent approaches propose refinement stages for RGBD data wit polarimetric cues~\cite{kadambi2017depth}.
CroMo goes one step further and researches cross-modal self-supervision with polarimetric and ToF cues on scenes with strong ambient light within their study~\cite{CroMo} which comprises a promising direction for future sensor fusion.
Their ground truth sensor is an active stereo intel RGBD camera.
As RGB+P data essentially provides RGB information with orthogonal polarimetric cues, we decided to include a polarisation sensor and two intel RGBD cameras (dToF and active stereo) on our rig and investigate monocular depth estimation pipelines trained on various sensor data.

\smallskip
\noindent
\textit{Synthetic Depth Renderings.}
In order to produce pixel-perfect ground truth, some scholars decided to render data for synthetic scenes~\cite{mayer2016large}.
While this produces the best possible depth maps, the scenes are artificially created and lack realism, causing pipelines trained on the Sintel~\cite{Butler:ECCV:2012} or SceneFlow~\cite{mayer2016large} to suffer from the synthetic-to-real domain gap.
In contrast to these ideas, we follow a hybrid approach to leverage the pixel-perfect synthetic rendering capabilities of modern 3D engines and adjust highly accurate 3D models to real captured scenes in indoor environments.

\subsection{Learning to Estimate Depth from a Single RGB Image}
The task of estimating depth from a single image is inherently ill-posed.
Due to the non-invertible nature of the projection operator, many 3D scenes might correspond to the same 2D image, creating uncertainties for this task. Geometric priors or knowledge of features representation encompassed in a neural network can be utilized to recover depth:

\smallskip
\noindent
\textit{Supervised monocular depth.}
Deep learning enabled this task for a broader set of real scenes.
Networks can learn to predict depth with supervised training. Eigen et al.~\cite{eigen2014depth} designed the first monocular depth estimation network by learning to predict coarse depth maps, which are then refined by a second network. Laina et al.~\cite{laina2016deeper} improved the latter model by using only convolutional layers in a single scale CNN.
The final network is lighter and thus requires far fewer data to be trained. Despite the improvements, the required ground truth depths limit the potential of these methods, often reducing applications to outdoor scenarios~\cite{Geiger_2013}.
A way of bypassing this limitation is to use synthetic data for depth estimation~\cite{mayer2018makes}.
The resulting domain difference has been tackled by Guo et al.~\cite{Guo_2018_ECCV} who found a way to reduce the gap between synthetic and real data and MiDaS~\cite{Ranftl2019} mixes dataset to enrich the learning signal with 3D movies, which results in a pipeline that generalizes better to unknown scenes.
To predict high-resolution depth infomation, most methods use multi-scale features or pose processing stages~\cite{Miangoleh2021Boosting,Wei2021CVPR} which complicate the training and increase the computational cost.
Fu et al.~\cite{fu2018deep} proposed a space increasing discretization of depth to turn the regression problem into an ordinal one leading to simplified training and giving sharper depth predictions.
If not trained on a massive set of data, the generalization capabilities of such methods are, however, limited.

\smallskip
\noindent
\textit{Self-supervised monocular depth.}
Unsupervised monocular methods have been investigated to bypass the limited available dataset with ground truth depths.
Xie et al.~\cite{xie2016deep3d} and Garg et al.~\cite{garg2016unsupervised} first proposed to use stereo images to train a network to predict depth as an intermediate output.
With depth, the left image is warped into the right one.
The consistency error between the predicted right image and the original one enables an unsupervised training of the network using photometric consistency.
Monodepth.~\cite{Godard_2017} added a left-right consistency loss to leverage left to right and right to left consistencies based on depth predictions from a single image.
Pillai et al.~\cite{pillai2019superdepth} used a fully differentiable flip augmentation layer to handle inconsistent boundaries depth predictions and added sub-pixel convolutional layers to generate higher-resolution depths from the usual low-resolution features.
If unsupervised stereo methods provide reasonable quality depth estimations, they still require synchronized images.
Monocular training methods have been developed for training networks with consecutive images from a single camera.
The pose variation between each frame and the depth are estimated simultaneously.
Unfortunately, this additional required prediction negatively affects the prediction performances.
Monodepth2~\cite{monodepth2} reduced the gap between the stereo and monocular training by automasking and a minimum reprojection loss.
Yang et al.~\cite{yang2018lego} leveraged the physical link between surface normals and depth to refine the depth. This idea was further improved by Yang et al.~\cite{yang2018unsupervised} by enforcing consistency with edges.
Other multi-task learning frameworks have been developed to use consistencies between depth and optical flow~\cite{Yin_2018} or semantic segmentation \cite{chen2019towards}.
Spencer et al.~\cite{spencer2020defeat} increased the robustness of the depth predictions, especially in challenging setups like night scenes, leveraging features consistency in the warping process.
Finally, Liu et al.~\cite{liu2019neural} proposed to predict simultaneously depths and their associated distribution, which are accumulated over time and integrated by a Bayesian filtering framework increasing the temporal stability and robustness of the final predictions.
The popularity of transformer architectures has also impacted depth estimation where attention volumes improved planar structures~\cite{huynh2020guiding}, and patch-wise attention was used for general depth estimates~\cite{lee2021patch}.

To make monocular depth also estimates suitable for augmented reality applications, some scholars proposed methods to make depth consistent over time.
Luo et al.~\cite{luo2020consistent} proposed a test-time refinement strategy, and a more efficient cost volume aggregation over a sequence of frames for consistency is used by ManyDepth~\cite{watson2021temporal}.
While initially there was a trade-off between accuracy and consistency which prevented methods from being successful in both directions, the spatio-temporal attention mechanism from Ruhkamp et al.~\cite{ruhkamp2021attention} closed this gap.

In order to compare the effect of various ground truth sensors for the task of monocular depth estimation, we utilized the ResNet backbone of the popular Monodepth2~\cite{monodepth2} together with its various training strategies and investigated core differences of the performance.
To produce highly reliable ground truth data, we follow the 6D pose labeling strategy proposed by PhoCal~\cite{PhoCal} which utilizes the forward kinematics of a robotic arm in gravity compensation mode to annotate objects with high precision.
\begin{figure*}[tbp]
 \centering
    \includegraphics[width=\linewidth]{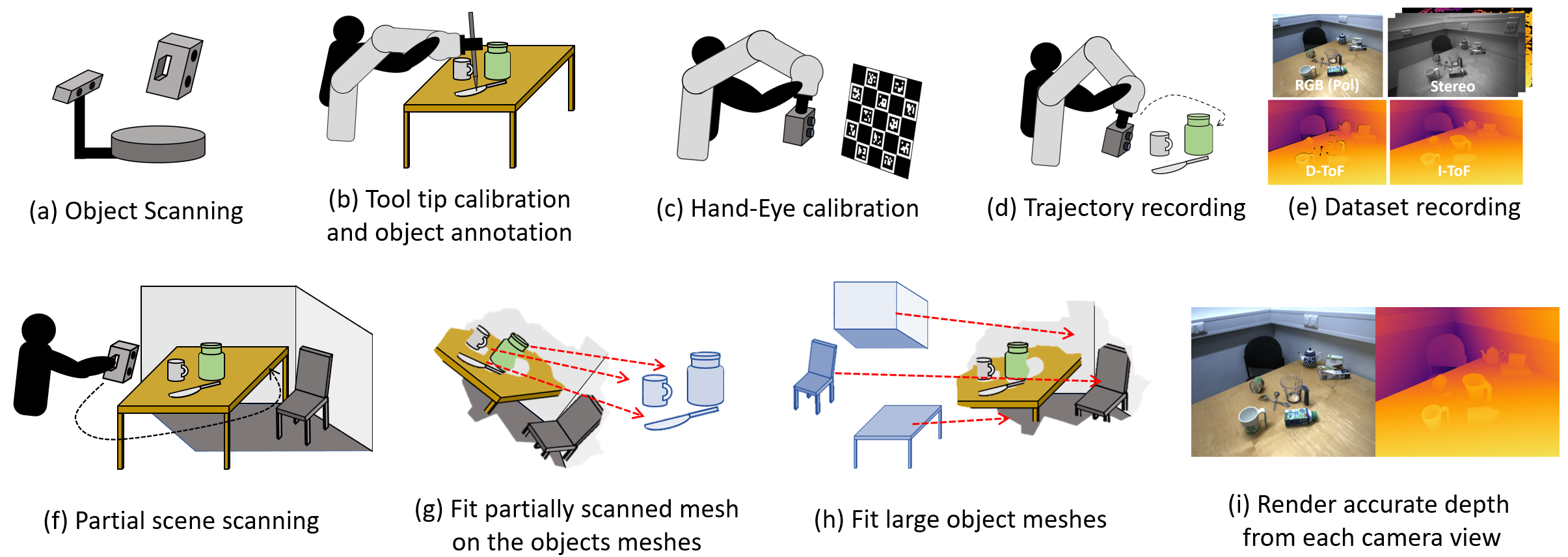}
    \caption{An overview of our dataset annotation. On the top of the dataset annotation pipeline using the robotic arm as in~\cite{PhoCal}, we add meshes of the background and large object and extra steps to annotate them to render the dense and accurate depth map.}
    \label{fig:annotation_overview}
\end{figure*}

\section{Dataset Acquisition}

\label{sec:dataset}
Similar to~\cite{PhoCal}, the scenes of our dataset are composed of multiple object classes with different shapes and materials, including the photometrically challenging ones, such as objects with reflective or transparent surfaces and synchronized multimodal sensor images as input. Different from~\cite{PhoCal}, it focuses more on the depth estimation problem, which provides depth maps in the larger scale of the room with different modalities sensor images and the high quality rendered depth from the annotated scene from the viewpoint of each sensor. In this section, we describe our dataset acquisition pipeline as shown in Fig.~\ref{fig:annotation_overview}.

 
\subsection{Object and Scene Scanning}
Unlike well known 3d or depth datasets, such as \cite{replica19arxiv,Matterport3D,dai2017scannet}, which instead scan the scene as a whole, every single object on our scenes, including chairs and backgrounds as well as small household objects, are scanned prior and then annotated in the scene to produce highly detailed and dense depth ground truth.

\begin{figure*}[htbp]
 \centering
    \includegraphics[width=\linewidth]{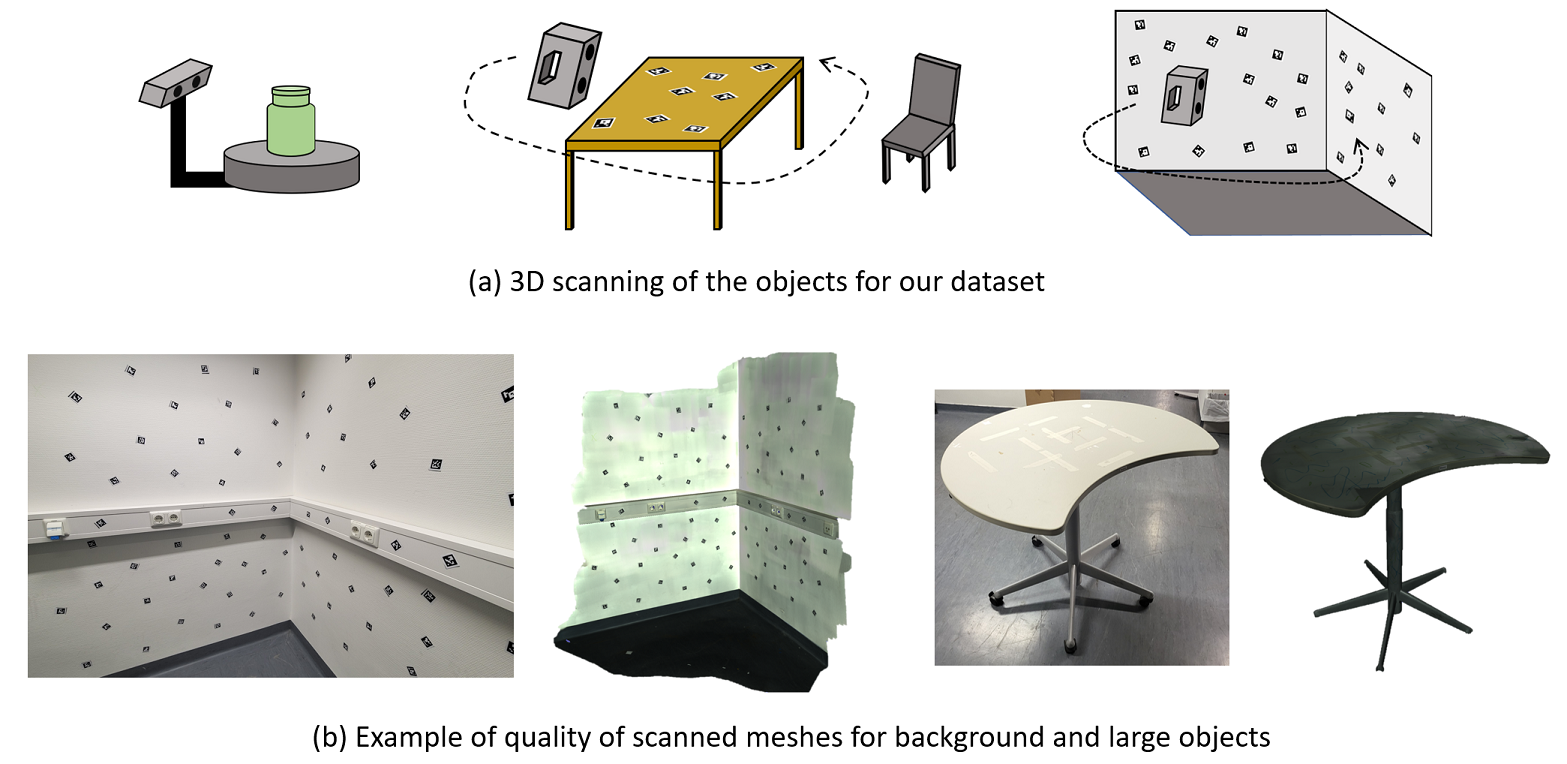}
    \caption{Two scanners are used for the scanning of the objects. One with the rotating table for the small objects ((a), left) and a hand-held scanner for the large objects and the backgrounds ((a), middle, right). (b) shows the quality of scanning on the background and the large object such as a table.}
    \label{fig:scanning}
\end{figure*}

As shown in Fig.~\ref{fig:scanning} two scanners are used for our dataset. The small objects are scanned by EinScan-SP 3D Scanner (SHINING 3D Tech. Co., Ltd., Hangzhou, China) as in~\cite{PhoCal} while large objects such as tables or chairs and backgrounds are scanned by handheld Artec Eva 3D Scanner (Artec 3D, Luxembourg). For the objects or the area with the challenging material self-vanishing 3D scanning spray (AESUB Blue, Aesub, Recklinghausen, Germany) is used to ease the scanning, and for a sizeable textureless area such as tables and walls, small markers are attached to the surface to help the 3D scanner localize.

\subsection{Scene Acquisition \& Sensor Setup}
Multiple combinations of objects are placed on the pre-scanned table within the corner of the pre-scanned background for each scene. KUKA LBR iiwa 7 R800 (KUKA Roboter GmbH, Germany) with a positional reproducibility of $\pm0.1$~mm is used for both annotation of the objects and the scene recording with the trajectories.

\begin{figure*}[!hb]
 \centering
    \includegraphics[width=0.4\textwidth]{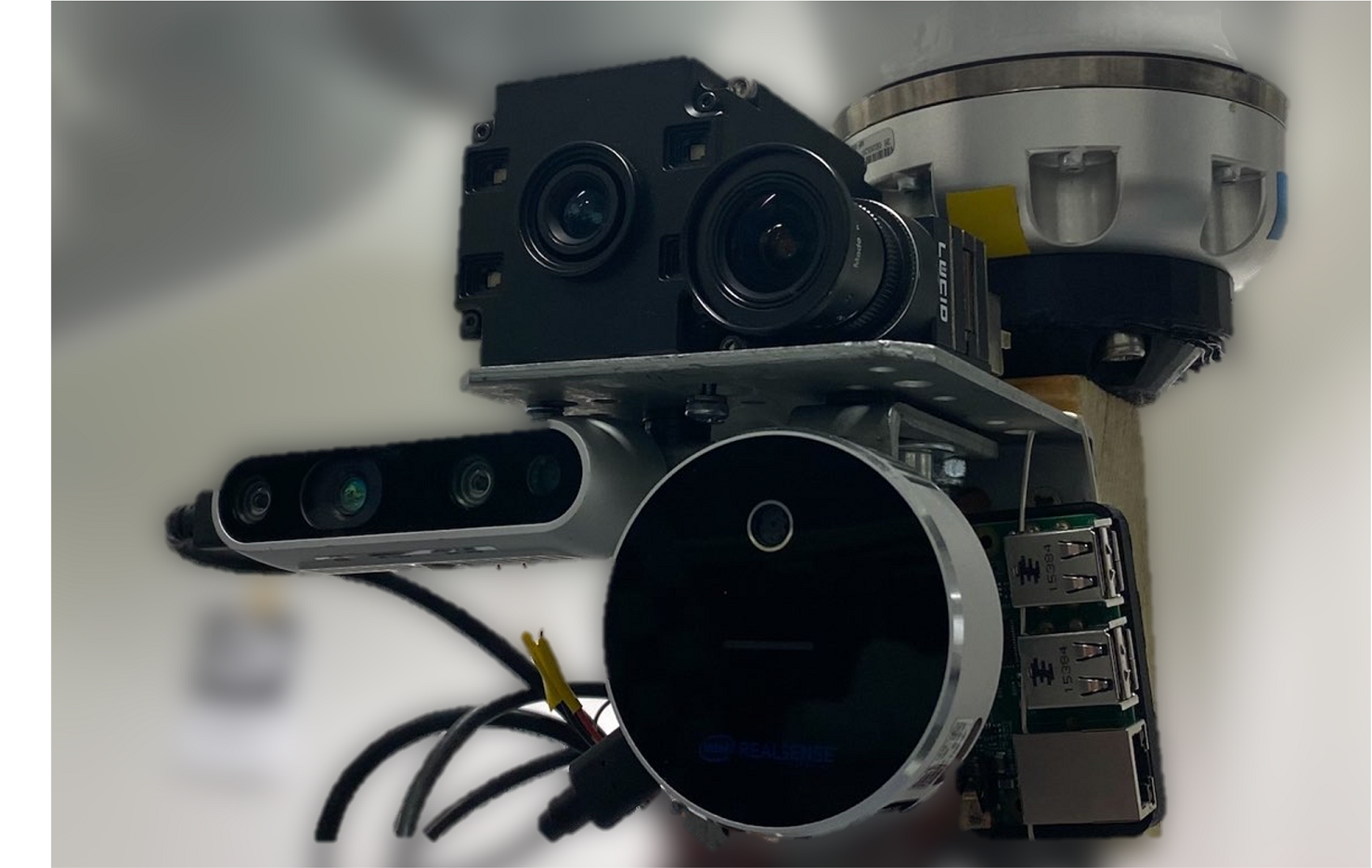}
    \caption{Hardware setup with all cameras mounted on a custom rig and rigidly attached to the robot end-effector. Sensors in order top-down-left-right: iToF, Polarimetric RGB, Active Stereo, dToF.}
    \label{fig:hardware}
\end{figure*}

Our dataset features a multi-modal setup with four different cameras, which provide four different types of input images (RGB, polarization, stereo, Indirect ToF (I-ToF) correlation) and three different depth images with different modalities (Direct ToF(D-ToF), I-ToF, Active Stereo). For the RGB and polarization image, we use a Phoenix 5.0 MP Polarization camera equipped with Sony IMX264MYR CMOS (Color) Polarsens (PHX050S1-QC, LUCID Vision Labs, Canada).
To acquire stereo images, we use Intel RealSense D435 (Intel, USA)  with the infra-red projector off. As for depth sensors, Intel RealSense L515 is used to produce D-ToF depth, Intel Realsense D435 is used for active stereo depth, and Lucid Helios ToF camera (HLS003S-001, LUCID Vision Labs, Canada) is used for the I-ToF Depth. 
 All cameras are fixed in the one custom rig designed to be mounted in the robot's end-effector (see Fig.~\ref{fig:hardware}).

For the synchronization between the cameras as well as the robot's pose, we adapted the approach from \cite{PhoCal}, which stops the robot within the trajectory and then captures the images to ensure the perfect synchronization among all cameras and robot pose. On top of that, we use Raspberry Pi (Raspberry Pi Foundation, United Kingdom) to produce different triggering times for each camera to remove the interference between the depth sensors as all the three sensors use infrared, which introduces artifacts to each other.

\begin{figure*}[htbp]
 \centering
    \includegraphics[width=\linewidth]{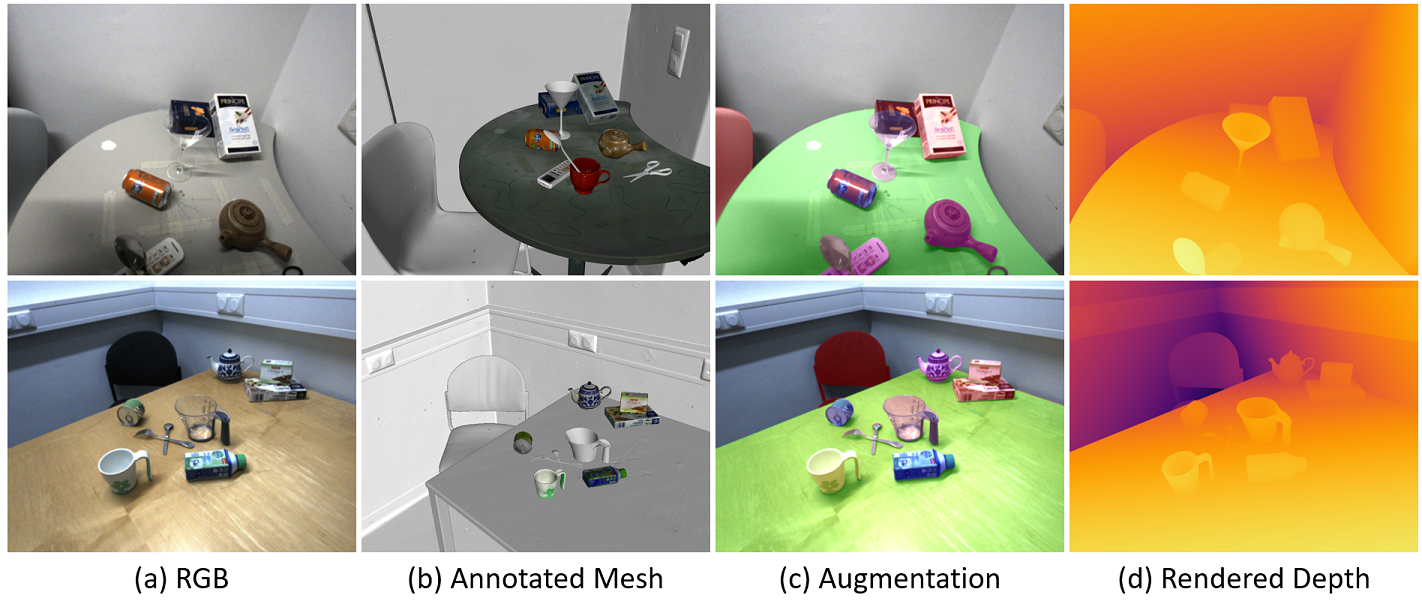}
    \caption{Example of the quality of object and background annotation in RGB frame (a). We show the quality of our annotation and its depth with the annotated mesh objects in the scene (b), their rendered augmentation mask on the image (c) and rendered depthmap of the scene (d). The augmentation mask shows that the all the objects are well aligned to the RGB image, which implies the quality of rendered depth where the occlusion boundary is well aligned in our dataset.}

    \label{fig:annotation_quality}
\end{figure*}

\subsection{Depth Quality}
\begin{figure*}[htbp]
 \centering
    \includegraphics[width=\linewidth]{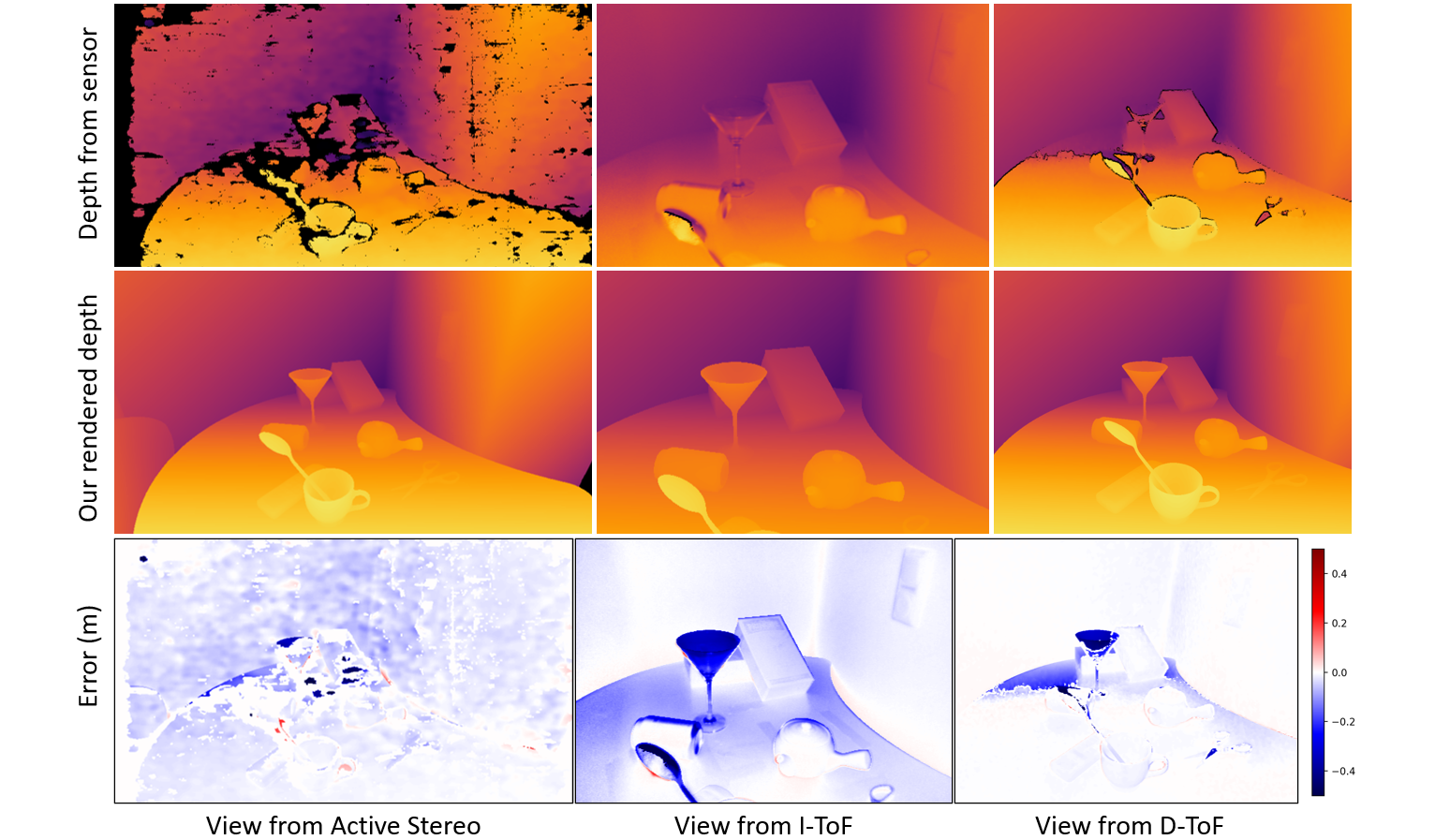}
    \caption{Depth quality comparison on different modality from scene in Fig.~\ref{fig:annotation_quality}. Each sensor acts different on differently depending on surface materials and normal of objects in the scene, which implies that choosing the depth camera is important aspect to have a good ground truth for the given situation.}
    \label{fig:depth_camera_quality}
\end{figure*}

For the small objects near the robot arm, custom made tooltip attached to the robot's end-effector (Fig.~\ref{fig:annotation_overview} (b)) is used to obtain objects' sparse point locations from the robot base, which can produce the annotation with an accuracy of 0.80mm after taking account all the source of error in pipeline~\cite{PhoCal} (i.e., from robot's noise to hand-eye calibration error). While for the large objects, the robot's limited working range made it impossible to annotate in this way. Hence, we use the handheld scanner to annotate the unreachable objects or background (Fig.~\ref{fig:annotation_overview} (f-h)). We first partially scan the scene and then fit the partial scan to annotated object mesh, then the meshes of unreachable objects or background are fitted into the partial scan. As this method is not as accurate as using the robot's tooltip as partial scanning is prone to noise as well as the mesh itself is incomplete, the pose of fitted meshes are later manually adjusted and inspected by humans until so that the rendering of the objects fit nicely on each camera's image globally. Fig.~\ref{fig:annotation_quality} shows the overall quality of our annotation by overlaying the rendered depth in each camera's image.
The 3D labeling process with the robotic manipulator has been reported with a point RMSE error of $0.80$~mm~\cite{PhoCal}.
For comparison, a Kinect Azure camera has a random depth-sensing error in the working range with a standard deviation of $17$~mm~\cite{liu2021stereobj}.
This accuracy difference allows for a reliable objective analysis of the depth error where marginal noise can be neglected as the error scales are more than one order of magnitude away from each other. We show the quality of depth map from other commercially available depth cameras in one example scene in Fig.~\ref{fig:depth_camera_quality} to show that the ground truth commonly used could not be proper. More comparisons on different scenes are attached in the supplementary material.

\subsection{Scene Statistics}

Total 7 different indoor areas, 6 different tables, and 4 chairs are scanned via the handheld scanner as the background and large objects, while total 64 household objects from 9 different categories of objects (bottle, can, cup, cutlery, glass, remote, teapot, tube, shoe) are scanned via rotating table scanner. Our dataset is composed of total 13 scenes, with 10 scenes for training and 3 scenes for testing. Each scene is recorded with 2 different trajectories of 2-300 frames with and without the objects, which sums up 4 trajectories and 800-1200 frames per scene, thus in total around 10k frames for the training set and around 3k frames for the test set. The test sets are designed with different setups to test the different aspects of the network's generalization. The three test scenes have different background setups: 1) Seen background, 2) Seen background with different lighting condition and 3) Unseen background and table, with different three different object setups: 1) Seen objects 2) Unseen objects from the seen category 3) Unseen objects from unseen categories (shoe and tube). Table~\ref{tab:dataset_comparison} compares our dataset with various existing setups. To the best of our knowledge, HAMMER is the only multi-modal dataset comprising RGB, ToF, Stereo, Active Stereo, and Polarisation modalities at the same time together with reliable ground truth depth maps. Detailed descriptions of the file formatting and structure of the dataset are in the supplementary material.

\begin{table}[tbp]
\setlength{\tabcolsep}{7pt}
\centering
\caption{\textbf{Comparison of Dataset Properties}. Shown are differences between HAMMER (Ours) and previous multi-modal depth datasets for indoor environments.}
\label{tab:dataset_comparison}
\begin{tabular}{r | c c c c c c c c c c c c } 
\toprule
  \small Dataset &
  \rotatebox[origin=c]{90}{\small RGB}          &
  \rotatebox[origin=c]{90}{\small Acc. GT}        &
  \rotatebox[origin=c]{90}{\small dToF/iToF}          &
  \rotatebox[origin=c]{90}{\small Stereo} &
  \rotatebox[origin=c]{90}{\small Active Stereo} &
  \rotatebox[origin=c]{90}{\small Polarisation} &
  \rotatebox[origin=c]{90}{\small Indoor}          &
  \rotatebox[origin=c]{90}{\small Real}         &
  \rotatebox[origin=c]{90}{\small Sequences}        &
  \rotatebox[origin=c]{90}{\small Frames}       \\ 
\midrule
Sturm~\cite{sturm2012benchmark}         & \checkmark & - & - & - & -          & -          & \checkmark & \checkmark & \checkmark & ${>}10k$   \\
Agresti~\cite{Agresti_2019_CVPR} & -          &  - & -/\checkmark & -          & -& -          & \checkmark & \checkmark & -          & $113$      \\
Guo~\cite{Guo_2018_ECCV}       & -          & \checkmark   & -/\checkmark & -          & - & -          & \checkmark & - & - & $2000$\\
Zhu~\cite{zhu2019depth}      & (\checkmark) & -            & -          & - & - & \checkmark & \checkmark & \checkmark & -        & $1$   \\
Qiu~\cite{qiu:2019a}             & \checkmark   & -            & -          & - & - & \checkmark & \checkmark & \checkmark & -          & $40$  \\
Ba~\cite{ba2020deep}            & \checkmark   & - & -          & - & - & \checkmark & \checkmark & \checkmark & -          & $300$ \\
Kadambi~\cite{kadambi2017depth}  & \checkmark   & - & -          & - & - & \checkmark & \checkmark & \checkmark & -          & $1$   \\
CroMo~\cite{CroMo} & - &  -  & -/\checkmark & \checkmark & \checkmark & \checkmark & (\checkmark) & \checkmark & \checkmark & ${>}10k$    \\
\textbf{Ours} & \checkmark & \checkmark & \checkmark/\checkmark & \checkmark & \checkmark & \checkmark & \checkmark & \checkmark & \checkmark & ${>}10k$\\
\bottomrule
\end{tabular}
\end{table}
\section{Bench-marking Methods}
\label{sec:methods}
The dataset described above allows for the first time for rigorous, in-depth analysis of different depth sensor modalities as well as a detailed quantitative evaluation of learning-based dense scene regression methods when trained with different modality supervision.
Here, we leverage widely adopted methods to predict depth from monocular and stereo images~\cite{monodepth2} in supervised, semi-supervised with ground-truth camera poses and fully self-supervised configurations.

More specifically, we first train a U-net style encoder-decoder architecture with a ResNet18 encoder and skip connections to regress dense depth~\cite{monodepth2} with different supervision depth modalities to study the influence and the characteristics of widely adopted and novel sensors.
Additionally, we also analyze whether complementary semi-supervision via information of the relative pose between monocular acquisitions of the moving camera can overcome sensor issues. We build upon the static scene assumption and formulate a loss that assumes that the image of a scene from one view can be reconstructed in another view, given accurate depth and pose estimates.
This assumption is also used to train in a fully self-supervised setting in different configurations (e.g., mono and mono+stereo) following~\cite{monodepth2}.
An extensive study of the results from training to yield detailed insights into the capabilities and limitations of each sensor.

\paragraph{Dense Supervision}
We use different depth modalities of the dataset to supervise the depth prediction of the four pyramid level outputs after upsampling to the original input resolution with:
\begin{align}
	 	L_D=\sum_{i=0}^{i=3}{|\widetilde{D}}_i-{D}|,
  \label{equ:soft-argmin}
\end{align}
where $D$ is the ground truth depth map and $\widetilde{D}_i$ the predicted depth at pyramid scale $i$.


\paragraph{Semi-Supervision}
For the semi-supervised training, the ground truth relative camera pose is leveraged to perform differentiable backwards warping with bilinear interpolation with the predicted depth estimate to formulate the photometric image reconstruction as in the self-supervised training~\cite{monodepth2} (see below). Here, we also enforce the smoothness loss detailed below.

\paragraph{Self-Supervision}
As depth and relative pose prediction between consecutive frames of a moving camera can be formulated as coupled optimization problem, we follow established methods to formulate a dense image reconstruction loss through projective geometric warping~\cite{monodepth2}. 
In this process some temporal image $I_{t^\prime}$ at time $t^\prime$ can be projectively transformed to the frame at time $t$ via:
\begin{align}
I_{t^\prime \to t} = I_{t^\prime}\Big\langle proj(D_t, T_{t \to t^\prime}, K) \Big\rangle,
\label{eqn:warp}
\end{align}
where $D_t$ is the predicted depth for frame $t$, $T_{t \to t^\prime}$ the relative camera pose transformation, and $K$ the camera intrinsics.
The photometric reconstruction error~\cite{monodepth2,watson2021temporal,ruhkamp2021attention} between image $I_x$ and $I_y$ given by
\begin{align}
    {E_{\text{pe}}}(I_x,I_y) = \alpha\tfrac{1-\text{SSIM}(I_x, I_y)}{2} + (1-\alpha) \left \Vert I_x-I_y \right \Vert_{1}
\label{eqn:photo}
\end{align}
is computed between the target frame $I_t$ and each source frame $I_s$ with $s \in S$ and the pixel-wise minimum error is retrieved.

$\mathcal{L}_{\text{photo}}$ is finally defined over $S \in \{t-10, t+10 \}$ as
\begin{align}
    \mathcal{L}_{\text{photo}} = \min_{s \in S}  {E_{\text{pe}}}(I_t,I_{s \rightarrow t}) .
\end{align}
Please note, that we sample $S$ with 10 frames offset due to small relative camera movement between frames and high frame rate.

The edge-aware smoothness is applied $\mathcal{L}_{\text{s}}$ as in previous works~\cite{monodepth2} to encourage locally smooth depth estimations with the mean-normalized inverse depth $\overline{d_{t}}$ as
\begin{align}
\mathcal{L}_{\text{s}} = \left|\partial_{x}\overline{d_{t}}\right|e^{-\left|\partial_{x}I_{t}\right|} + \left|\partial_{y}\overline{d_{t}}\right|e^{-\left|\partial_{y}I_{t}\right|}.
\end{align}

The final training loss for the self-supervised setup is defined with $\lambda_{\text{s}}=10^{-3}$ as:
\begin{align}
    \mathcal{L}_{\text{self-supervised}} = \mathcal{L}_{\text{photo}} +  \lambda_{\text{s}} \cdot \mathcal{L}_{\text{s}},
\end{align}


\subsection{Implementation Details}
We implement all models in PyTorch~\cite{paszke2017automatic} and train for 20 epochs for comparability using Adam~\cite{kingma2014adam}.
Monocular approaches are trained with a batch size of 12 on one NVIDIA RTX-3090 GPU. 
The RGB inputs are scaled to $480 \times 320$ for supervised training and to $320 \times 160$ for self-supervised training, respectively. 
The depth network regresses dense depth predictions on 4 pyramid levels, each half in resolution of the previous~\cite{monodepth2}.
In the self-supervised setup, we regress the relative pose between consecutive camera frames with a ResNet18 pose network according to~\cite{monodepth2}.
We choose an initial learning rate of $1 \times 10^{-4}$ for 15 epochs, which we decrease to $1 \times 10^{-5}$ after 15 epochs in the self-supervised setting.
For the supervised case, we start with a learning rate of $1 \times 10^{-3}$, which we decrease every 5 epochs by a factor of 10.
We perform the same color augmentations as~\cite{monodepth2}
(brightness$\pm0.2$, contrast$\pm0.2$, saturation$\pm0.2$, hue$\pm0.1$). 
Augmentations are applied to the network input, while for the image reconstruction loss of the self-supervised method the original images are used.
\section{Experiments \& Evaluation}
\label{sec:experiments}

The pipeline described in section~\ref{sec:methods} is trained on various alleged ground truth data with full supervision, and in self- and semi-supervised setups.
Results are reported in tables~\ref{tab:depth_supervision_results} and~\ref{tab:trainedon_vs_tested_on}.

Table~\ref{tab:depth_supervision_results} shows the results for training with various types of commonly used depth sensor data.
While it can be clearly seen (upper part) that the GT training provides the most reliable results in terms of RMSE overall, it is quite interesting that the error varies from background to object. This can be explained due to the fact that the photometric complexity of the objects varies and that the objects pose on average more challenges due to their material properties. Not surprisingly, the training with ToF sensors suffers particularly when the photometric complexity is high (e.g., for reflective and transparent objects) where the structured light sensor is capable to project some patterns onto diffusely reflective surfaces.
Interestingly, the self- and semi-supervised setups help to recover information in these challenging setups to some extent, such that these cases even outperform the ToF supervision for photometrically challenging objects. In contrast, more superficial structures (such as the background) benefit from the ToF supervision.

In table~\ref{tab:trainedon_vs_tested_on}, we analyze what it means to evaluate alleged ground truth sensors without taking their advantages and drawbacks into account.
For this, we utilize our pipeline trained on every single modality, namely iToF, dToF, active stereo (with pattern projector), and compare the evaluation on its own data.
It should be noted that such an evaluation is the typical case when a ground truth sensor is used without questioning its advantages or drawbacks for a specific material or scene.
In particular, in comparison with table~\ref{tab:depth_supervision_results}, it is clearly visible that the models learn the characteristics of the individual sensors and overfit to its depth errors since the quantitative evaluation is significantly lower on its own data.
This result, however, hides the ground truth characteristics, ultimately leading to an incorrect assumption that the pipeline works significantly better than it actually does.
The lower RMSE on the modality compared to high RMSE on the ground truth is significant, especially for the ToF sensors on complex photometric objects where the difference, e.g., for dToF with $117.4$~mm RMSE is more than the assumed error itself.

\begin{table*}[htbp]
\centering
\caption{\textbf{Depth prediction comparison when training with different modalities. }(Top) Evaluation against GT of depth predictions on the test set with dense supervision from different depth modalities. (Bottom) Evaluation of semi-supervised (pose GT) and self-supervised (mono and mono+stereo) training. The full scene (Full), the scene background (BG) and objects only (Obj) are evaluated. The objects are further split into textured, reflective and transparent. \textbf{best} and \underline{2nd best}. Error is reported as RMSE in mm.
}
\footnotesize
\resizebox{0.6\textwidth}{!}{
\begin{tabular}{ll|c|cc|ccc}
\shline
& &\multicolumn{1}{c}{Full} & BG & \multicolumn{1}{c}{Obj}  & Text. & Refl. & Transp. \\
\shline
\multirow{4}{*}{\rotatebox[origin=c]{90}{Supervised}}     & iToF & 113.29 & 111.13 & 119.72 & 54.45 & 87.84 & 207.89 \\
                                & dToF & 77.97 & \underline{69.87}& 112.83& \underline{37.88} & 71.59 & 207.85 \\
                                & Active Stereo & \underline{72.20} & 71.94& \underline{61.13}& 50.90 & \underline{52.43} & \underline{87.24} \\
                                & GT & \textbf{20.70} & \textbf{17.13} & \textbf{34.08} & \textbf{27.94} & \textbf{36.75}  & \textbf{44.53} \\
                                
\shline\\
\shline
\multirow{3}{*}{\rotatebox[origin=c]{90}{Sel/Sem}}     & Pose & \textbf{154.87} & \textbf{158.67} & \textbf{65.42}& \textbf{57.22} & \textbf{37.78} & \underline{61.86} \\
& M & 180.34 & 183.65& 85.51 & 84.26 & \underline{48.80} & \textbf{49.62} \\
                                & M+S & \underline{159.80} & \underline{161.65} & \underline{82.16} &\underline{71.24} & 63.92 & 66.48\\
\shline
\end{tabular}
}
\label{tab:depth_supervision_results}
\end{table*}
\begin{table*}[htbp]
\centering
\caption{\textbf{Depth modality over-fitting. }Evaluation of dense depth predictions when trained and tested on the identical modality. The full scene (Full), the scene background (BG) and objects only (Obj) are evaluated. The objects are further split into textured, reflective and transparent. Error is reported as RMSE in mm.
}
\footnotesize
\resizebox{0.5\textwidth}{!}{
\begin{tabular}{l|l|c|c|c}
\shline
\multicolumn{2}{l|}{Trained} & \multicolumn{1}{c|}{iToF} & \multicolumn{1}{c|}{dToF} &  \multicolumn{1}{c|}{Active Stereo} \\
\hline

\multirow{6}{*}{Tested on modality}     & Full & 50.57 & 
49.04 & 
87.44    
\\
\cline{2-5}
& BG & 49.83 & 
46.13 &  
86.80 
\\
& Obj & 50.01 & 
62.22 &  
85.05 
\\
\cline{2-5}
& Text. & 32.26 & 
47.47 &  
85.22 
\\
& Refl. & 93.75 & 
87.12 &  
72.41 
\\
& Transp. & 
56.34 & 
90.45 &  
90.77 
\\
\shline
\end{tabular}}

\label{tab:trainedon_vs_tested_on}
\end{table*}

\section{Discussion \& Conclusion}
\label{sec:limitation}
Although our dataset tries to provide scenes with varying backgrounds, the possible location of the scene is limited to corner of the room due to the limited working range of the robot.
It has to be noted that the annotation process for the background is tedious as extra steps with human annotation and inspection are required, which makes dataset hard to produce as large scale. The possible future work could be to investigate a more flexible way to annotate the objects and to track the camera, e.g. with a highly accurate tracking system comprising optical markers.
Aside of our investigations, we strongly believe that our dataset can stimulate further investigations for cross-modal fusion pipelines.

\clearpage
%
%
\bibliographystyle{splncs04}
\bibliography{literature}

\title{Is my Depth Ground-Truth Good Enough? HAMMER Dataset -- Supplementary Material}


\titlerunning{HAMMER dataset}
%
\author{HyunJun Jung\inst{1} \and
Patrick Ruhkamp\inst{1} \and
Guangyao Zhai\inst{1} \and
Nikolas Brasch\inst{1} \and
Yitong Li\inst{1} \and
Yannick Verdie\inst{2} \and
Jifei Song\inst{2} \and
Yiren Zhou\inst{2} \and
Anil Armagan\inst{2} \and
Slobodan Ilic\inst{1,3} \and
Ales Leonardis\inst{2} \and
Benjamin Busam\inst{1}
\\ \footnotesize{\fontfamily{qcr}\selectfont
hyunjun.jung@tum.de, b.busam@tum.de
}
}

\authorrunning{HJ. Jung et al.}
%
\institute{Technical University of Munich\and Huawei Noah's Ark Lab, \ \ \ \inst{3} \ Siemens AG}

\maketitle

\section{Detailed Background and Objects Description}
\label{sec:bckgr_obj_description}

\begin{figure*}[!b]
 \centering
    \includegraphics[width=\linewidth]{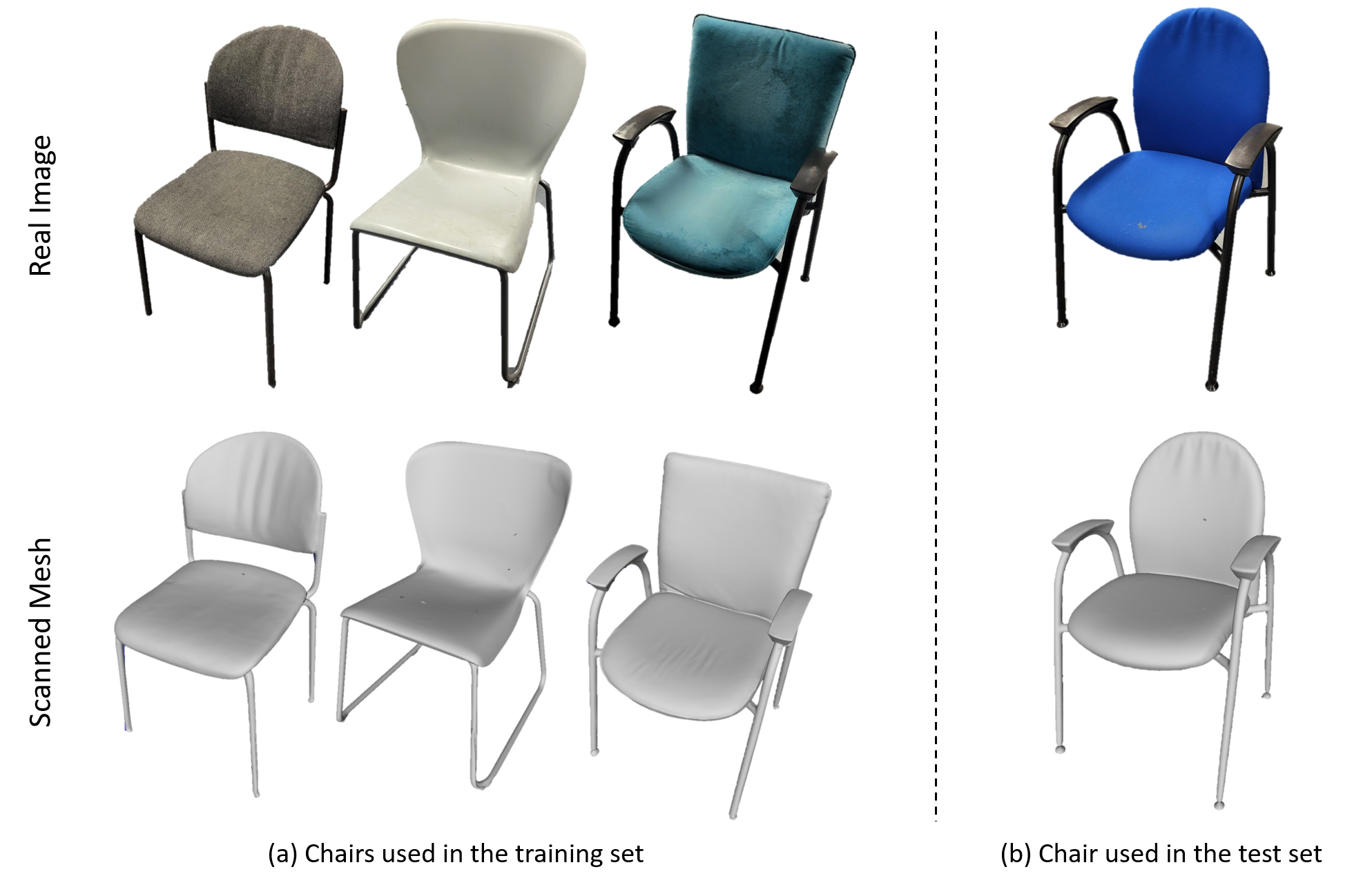}
    \caption{Chairs used in the dataset. Chairs in group (a) are used for the training set and the chair in (b) is used for the test set.}
    \label{fig:chair}
\end{figure*} 

\begin{figure*}[!p]
 \centering
    \includegraphics[width=\linewidth]{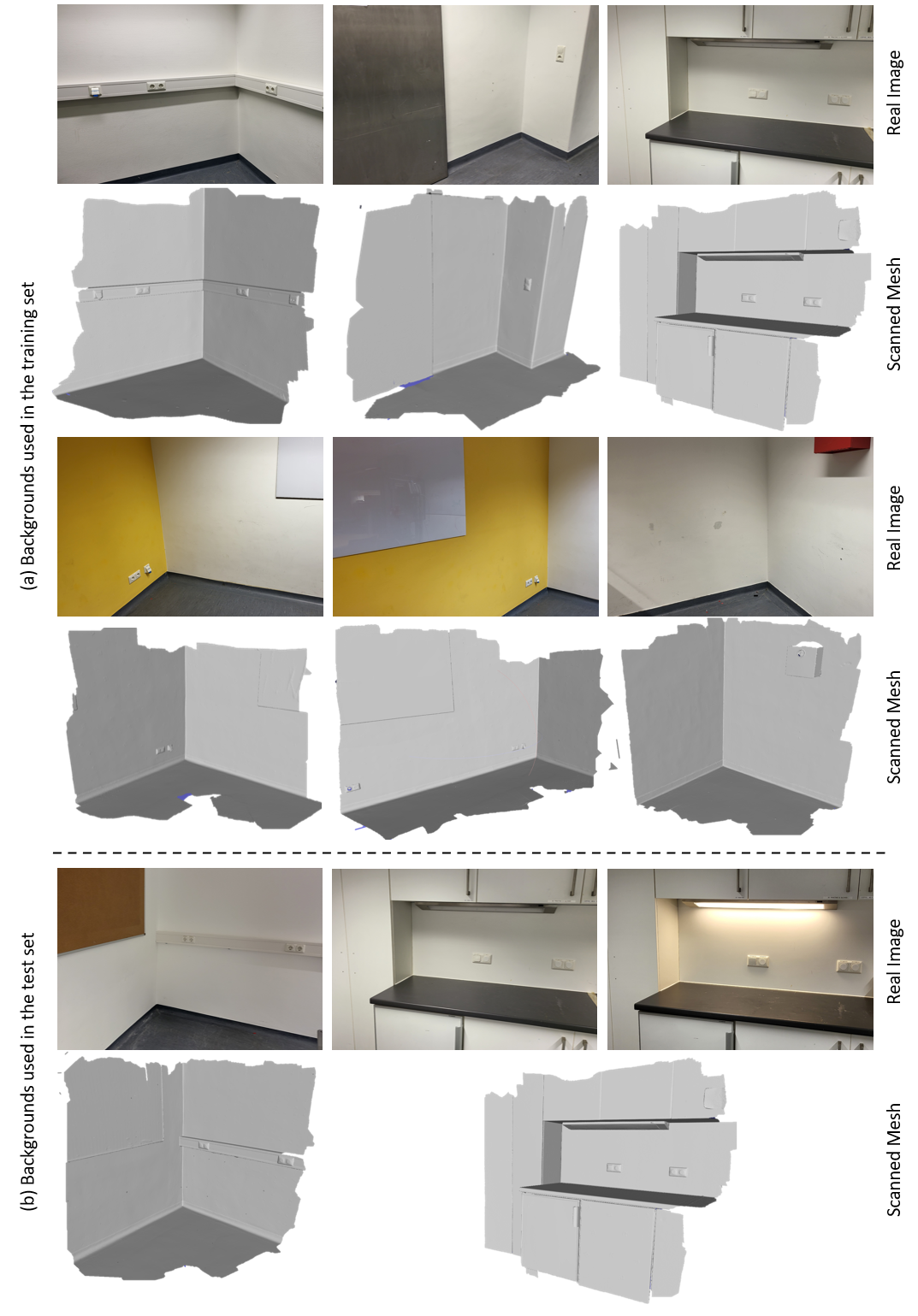}
    \caption{Backgrounds used in the dataset. Note that one of the background in the group (b) is also included in the training set, but we varied the lighting condition to provide different various factors for evaluation.}
    \label{fig:wall}
\end{figure*}

\begin{figure*}[!t]
 \centering
    \includegraphics[width=\linewidth]{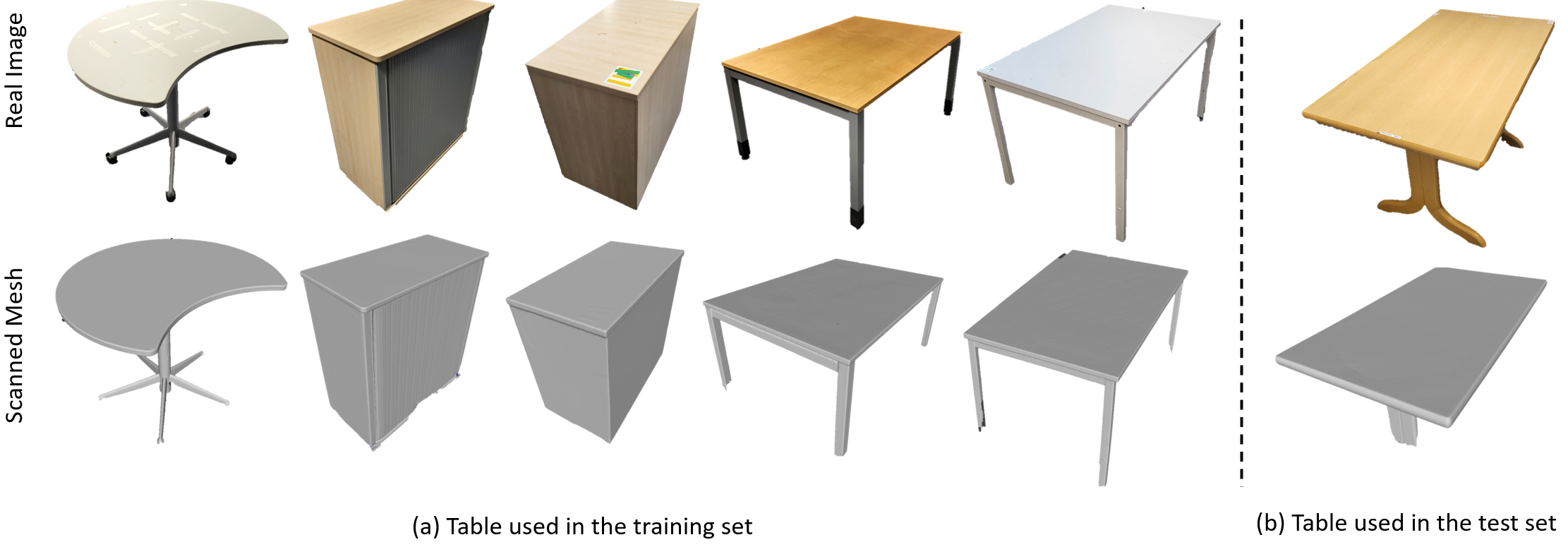}
    \caption{Tables used in the dataset. Tables in group (a) are used for the training set and the table in (b) is used for the test set. Note that, unlike small objects or chairs, we decide not to scan some parts of the large tables (e.g. end of their legs) as the cameras cannot see the part in their trajectories.}
    \label{fig:table}
\end{figure*}

\begin{figure*}[!b]
 \centering
    \includegraphics[width=\linewidth]{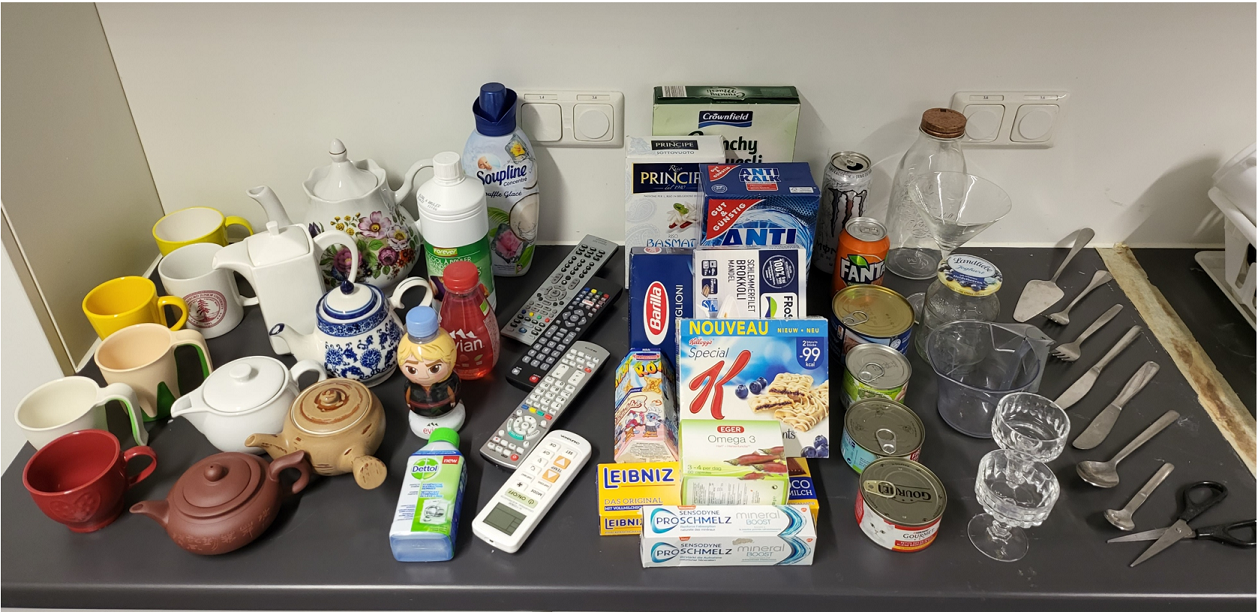}
    \caption{Collection of small household objects used in the training set. Objects from 8 household categories are used in the training set, 3 of which have photometrically challenging surface material - partially reflective (can), transparent (glass/plastic), reflective (cutlery).}
    \label{fig:obj_train}
\end{figure*}

\begin{figure*}[!t]
 \centering
    \includegraphics[width=\linewidth]{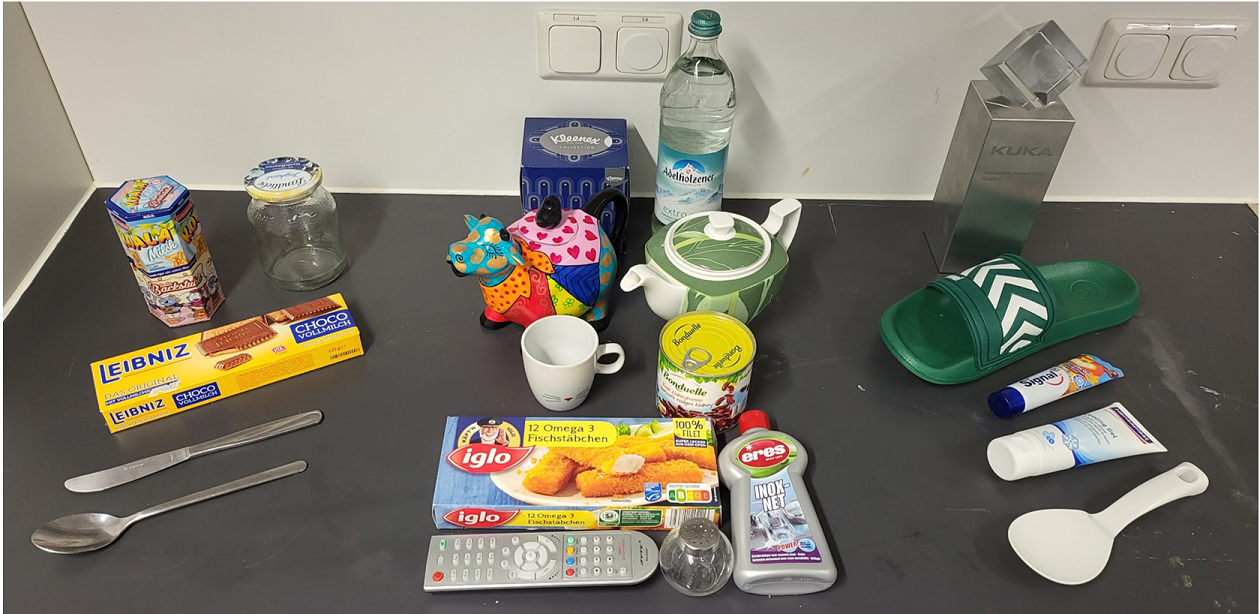}
    \caption{Collection of small household objects used in the test set. The test set comprises a mixture of seen (left column) and unseen (mid column) objects from 8 seen categories and a few objects from unseen categories (right column - tube, slipper, plastic kitchenware, trophy) are used.}
    \label{fig:obj_test}
\end{figure*}

As described in Sec.~4 in the main paper, our dataset comprises a total of 13 scenes divided into 10 scenes for training and 3 testing scenes composed of a mixture of 4 different chairs, 6 different tables, 64 household objects from 8 plus 4 different categories (i.e. cup, teapot, bottle, remote, boxes, can, glass, cutlery and tube, shoe, plastic kitchenware, trophy) and and 7 different indoor areas. Test sets have 1 unseen background and 2 seen backgrounds with and without different lighting and contain a mixture of seen/unseen objects from seen/unseen categories. In this section, we show detailed images of backgrounds, chairs, tables, and other objects. Fig.~\ref{fig:chair} and~\ref{fig:table} respectively show images of 3 chairs and 6 tables used in the dataset and their corresponding meshes. Fig.~\ref{fig:obj_train} and~\ref{fig:obj_test} show a collection of household objects used in training and test set. Fig.~\ref{fig:wall} shows 9 backgrounds used in the dataset and their corresponding meshes.

\section{Detailed Scene Description}
\label{sec:scene_description}

\begin{figure*}[!htbp]
 \centering
    \includegraphics[width=\linewidth]{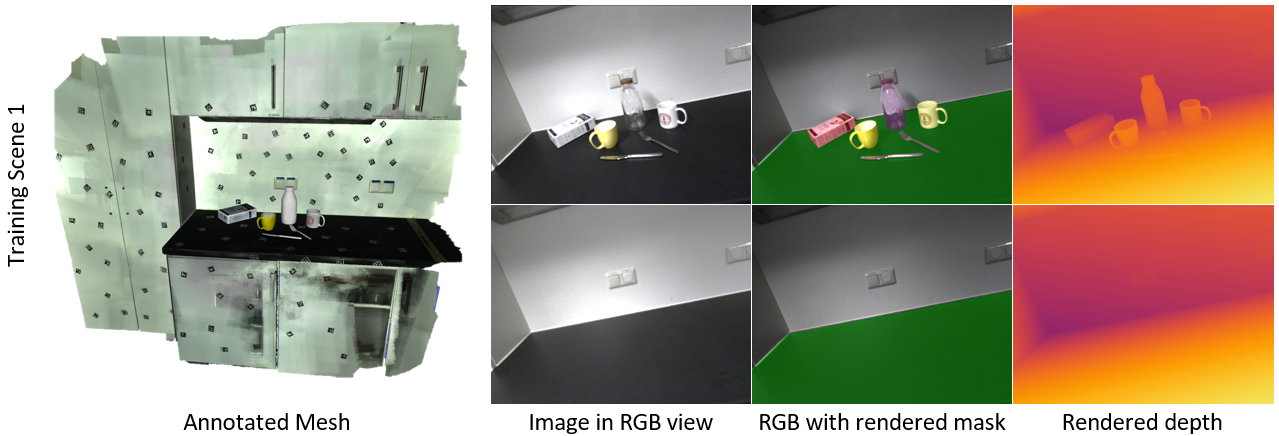}
    \caption{Example images from Training Scene 1. The annotated mesh is shown on the left together with an RGB view from the scene (second from left) with and without objects. The overlayed masks (second from right) and the rendered depth (right) illustrate the annotation quality of our data.}
    \label{fig:wall1}
\end{figure*}

\begin{figure*}[!htbp]
 \centering
    \includegraphics[width=\linewidth]{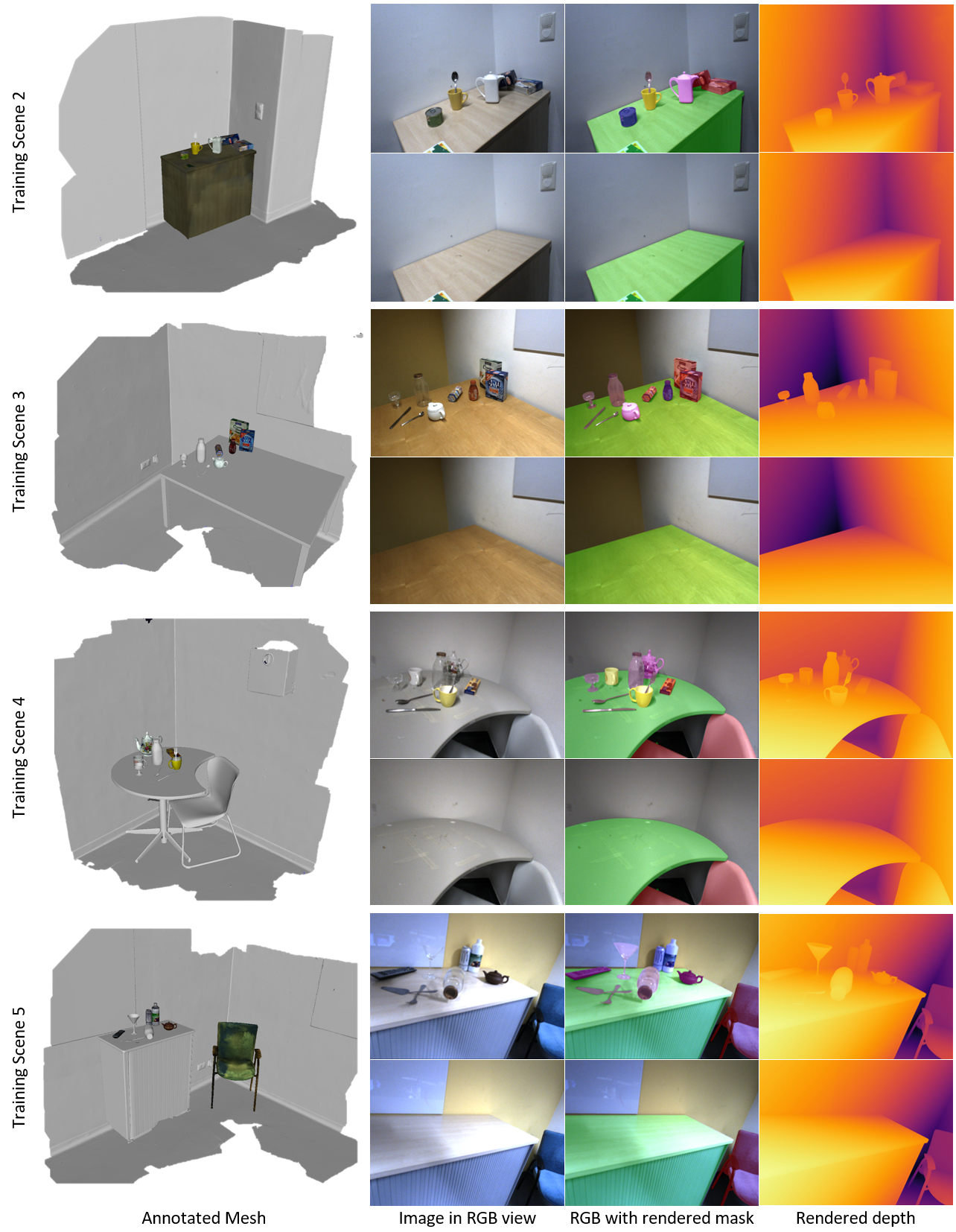}
    \caption{Example images from Training Scene 2-5. The annotated mesh for 4 different scenes is shown on the left together with an RGB view from the scene (second from left) with and without objects. The overlayed masks (second from right) and the rendered depth (right) illustrate the annotation quality of our data.}
    \label{fig:wall2}
\end{figure*}

\begin{figure*}[!htbp]
 \centering
    \includegraphics[width=\linewidth]{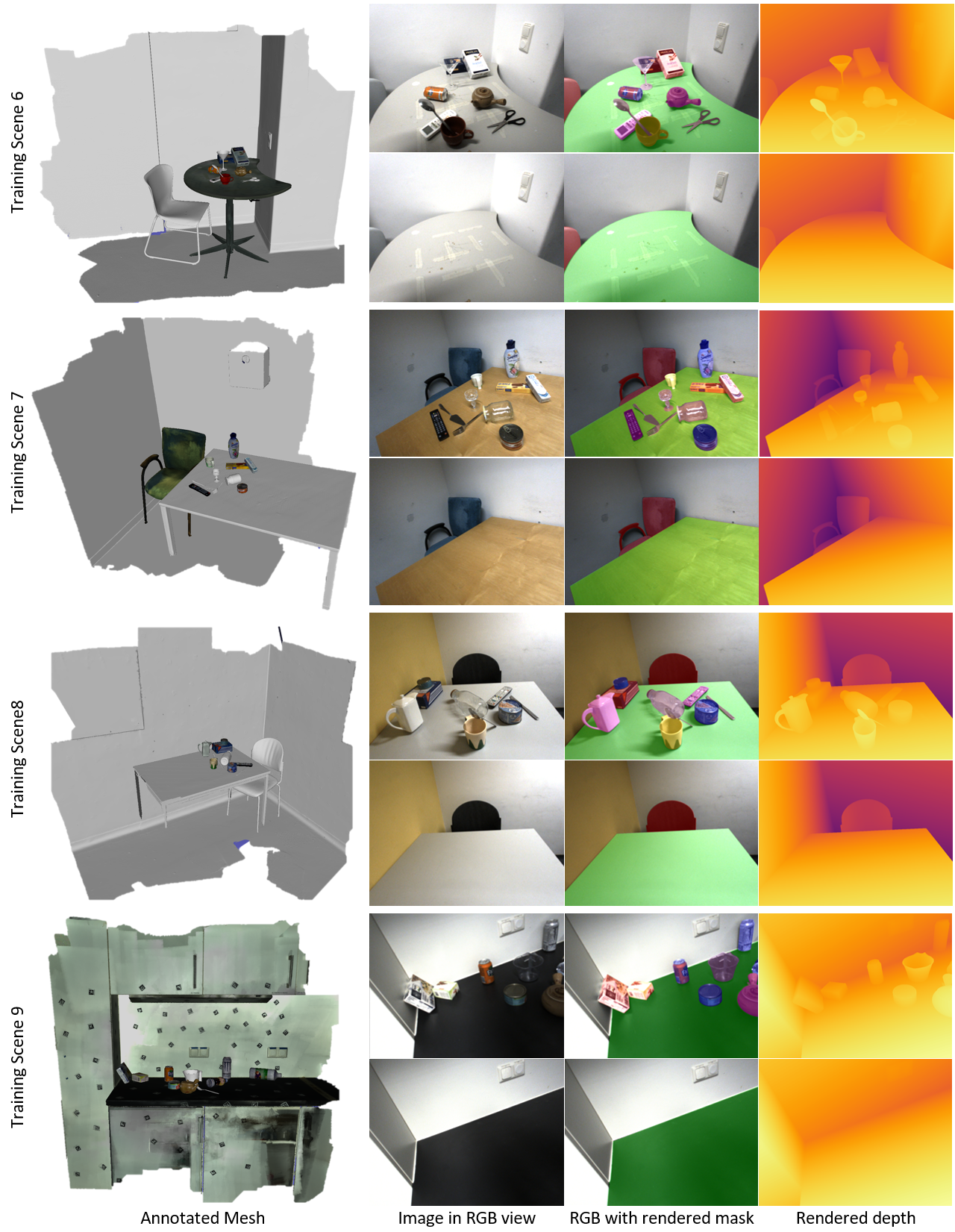}
    \caption{Example images from Training Scene 6-9. The annotated mesh for four different scenes is shown on the left together with an RGB view from the scene (second from left) with and without objects. The overlayed masks (second from right) and the rendered depth (right) illustrate the annotation quality of our data.}
    \label{fig:wall3}
\end{figure*}

\begin{figure*}[!htbp]
 \centering
    \includegraphics[width=\linewidth]{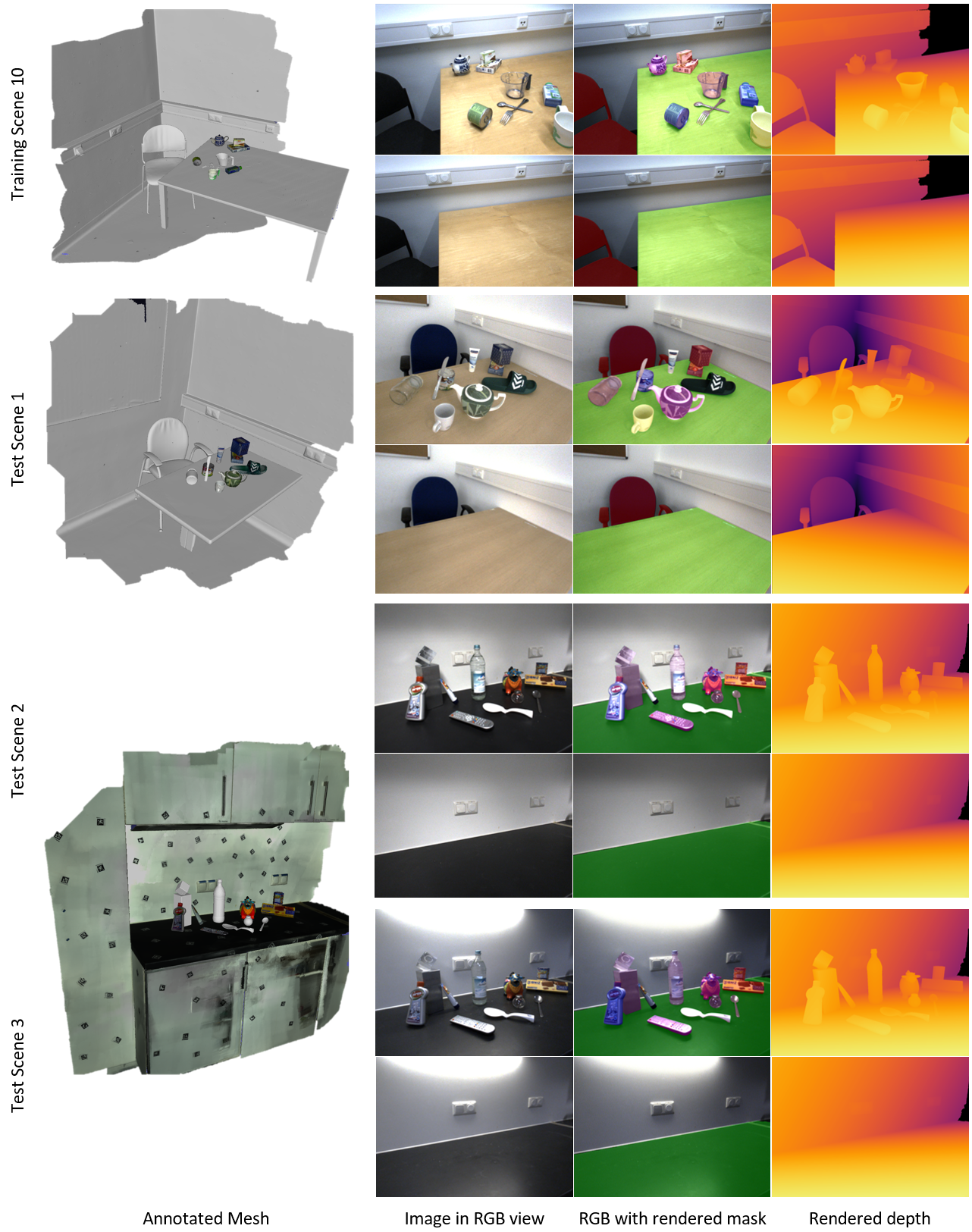}
    \caption{Example images from Training Scene 10 and Test scene 1-3. The annotated mesh is shown on the left together with an RGB view from the scene (second from left) with and without objects. The overlayed masks (second from right) and the rendered depth (right) illustrate the annotation quality of our data. Note that the test scene 2,3 are recorded in the exactly same pose and trajectory but with the different lighting.}
    \label{fig:wall4}
\end{figure*}

As described, our training set is composed of 10 scenes, and the test set is composed of 3 scenes. For each scene, we include 2 different trajectories. Each trajectory covers 2 setups with and without objects (naked scene). This sums up to 800-1200 frames per scene and a total of ca.~10k frames. In this section, we show several sample images of the scenes in Fig.~\ref{fig:wall1},~\ref{fig:wall2}, and~\ref{fig:wall3}, ~\ref{fig:wall4}. Each of them consists of an annotated mesh and RGB images with different types of rendering, which show the diversity and quality of HAMMER.

\section{Partial Scanning of the Scene and Mesh Fitting}
\label{sec:partial_scanning}

\begin{figure*}[!htbp]
 \centering
    \includegraphics[width=\linewidth]{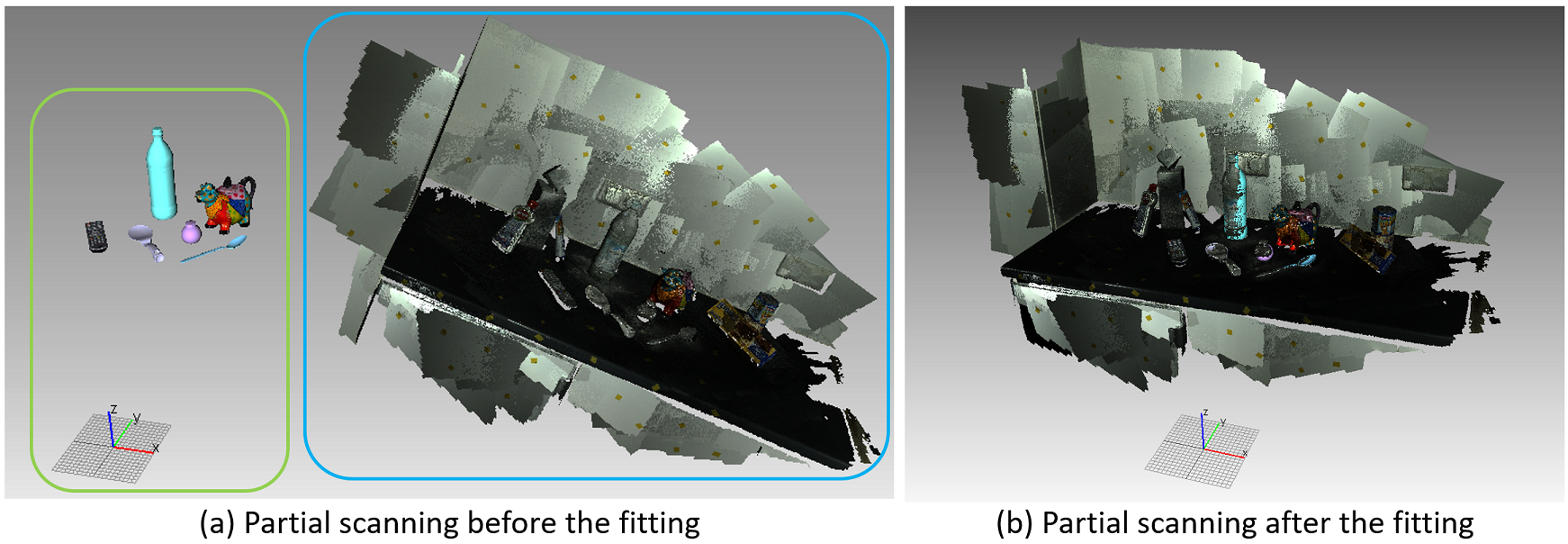}
    \caption{Example of partial scanning of the scene before and after the fitting on scene 13. Note that the center of the partial scanned mesh is aligned to robot base (xyz coordinate marker) after fitting it onto the mesh of the annotated objects.}
    \label{fig:annotated_objects}
\end{figure*}

\begin{figure*}[!htbp]
 \centering
    \includegraphics[width=\linewidth]{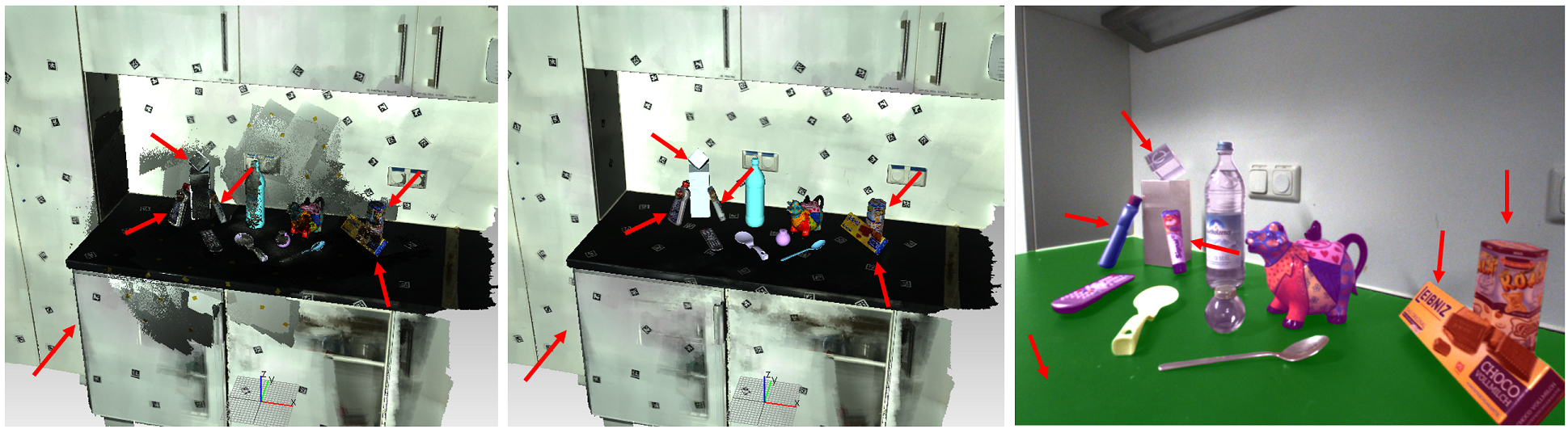}
    \caption{Example of far objects and background fitting onto partially scanned mesh. Left: Background and objects are fit to partial scans. Centre: All annotated meshes are shown without partial scans. Right: Corresponding scene from the camera viewpoint with augmented object masks. Note that the annotation quality of meshes with partial scans and robot arm is similar. The annotated meshes via partial scanning are marked with red arrows.}
    \label{fig:fitting_final}
\end{figure*}

As mentioned in Sec.~3 in the main paper, we use partial scanning and mesh fitting to annotate background, large objects, and objects outside the robotic workspace. This section shows images of partial scanning and the mesh fitting from one of the scenes as an example. The green box in Fig.~\ref{fig:annotated_objects}, (a) shows annotated meshes of the objects by the robotic arm. Once the objects are annotated, the scene is partially scanned with multiple viewpoints to make the scanning dense and cover multiple facets of the background. Note that the center of the scanning is not yet in the robot base coordinates (Fig.~\ref{fig:annotated_objects}, (a) blue box). Once the partial scanning is done, the scanned mesh is then fit onto the annotated objects, such that the partially scanned mesh origin concides with the robot base (Fig.~\ref{fig:annotated_objects}, (b)). Once the scanned mesh is put to robot base coordinates, we fit background, large objects, and distant objects meshes also in robot base coordinates to annotate them (Fig.~\ref{fig:fitting_final}, (a)). Fig.~\ref{fig:fitting_final}, (b-c) shows the result of the annotated mesh. 
For the mesh fitting, we used Artec Studio 10 Professional (Artec 3D, Luxembourg) which runs a point correspondence and ICP-based method to fit the meshes.

\section{Detailed Dataset Description}
\label{sec:detailed_dataset_description}

\begin{figure*}[!htbp]
 \centering
    \includegraphics[width=\linewidth]{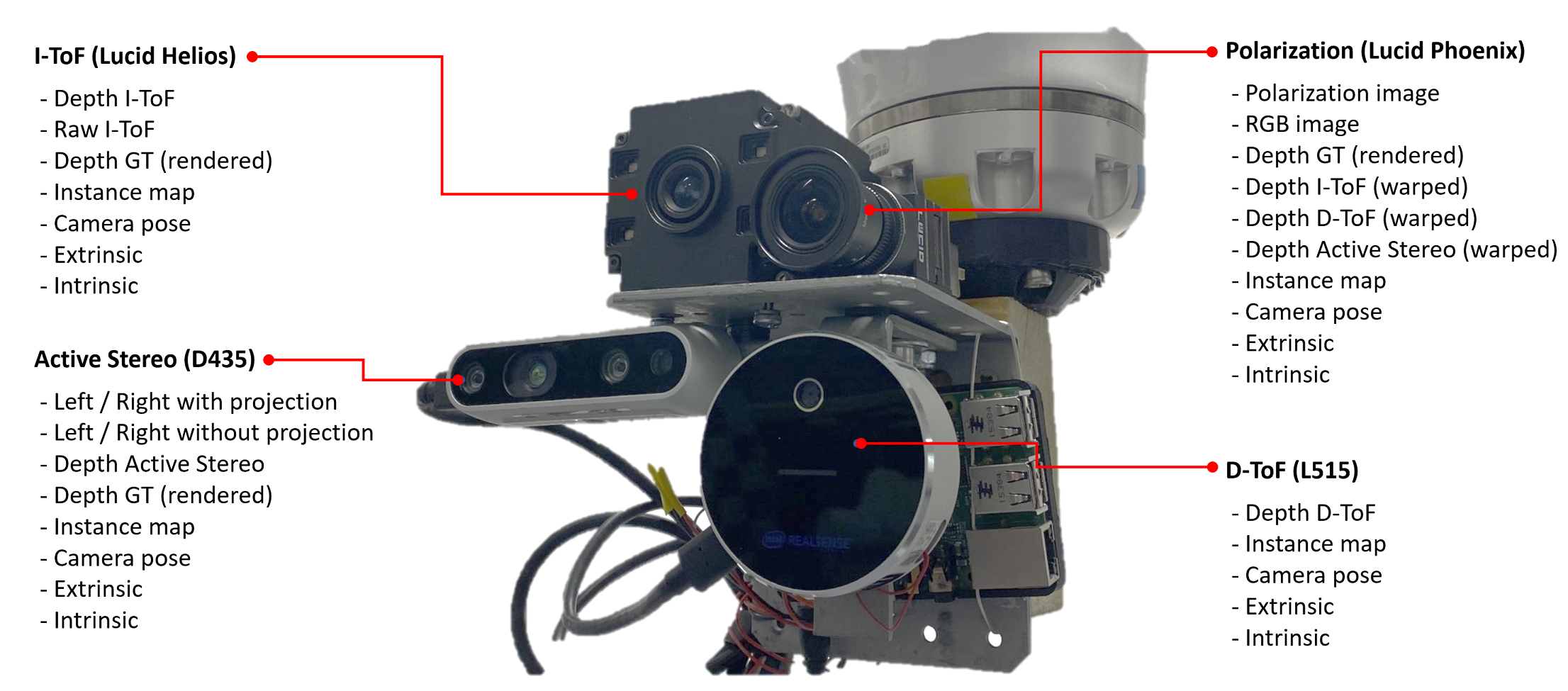}
    \caption{Lists of files included per sensor in our rig which comprises I-ToF, D-ToF, and Active Stereo depth sensors as well as a RGB+P polarization camera.}
    \label{fig:provided}
\end{figure*}

Sec.~3 of the main paper mentioned that our dataset uses multiple images/depth sensors to collect the dataset with highly accurate annotations of the scene using the robotic arm in a synchronized manner. This section shows the detailed description of files we include in our dataset. Fig.~\ref{fig:provided} shows the cameras in our rig and lists of files included per camera. Detailed descriptions and examples of the elements of the lists are shown in the following subsections.

\subsection{Polarization Camera}
\label{subsec:polarization_camera}

\begin{figure*}[!htbp]
 \centering
    \includegraphics[width=\linewidth]{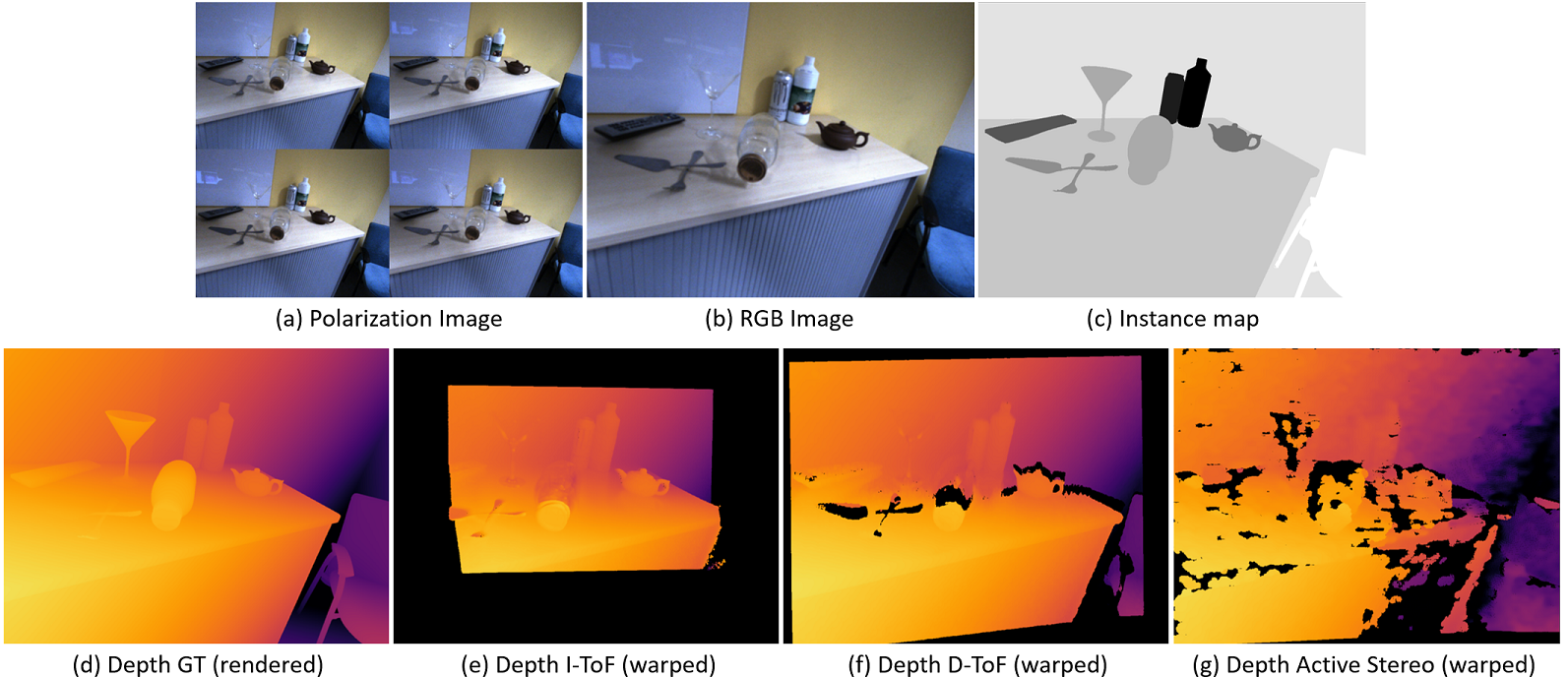}
    \caption{Example of the images included for the polarization camera input (top) together with instance label map and depth estimates warped onto the same coordinate reference frame.}
    \label{fig:polarization_included}
\end{figure*}

Fig.~\ref{fig:polarization_included} shows examples of images included for the polarization camera. As mentioned in Sec.~2 of the main paper, a polarization camera provides images with different polarization angles, which can extract cues like the surface normal by using the physical property of object material in the scene. The polarization camera we used in our dataset (See Sec.~3 in the main paper) provides polarized images at 4 different angles (0, 90, 180 270 degrees) which are saved in a single 2x2 image (Fig.~\ref{fig:polarization_included}, (a)). A regular RGB image is obtained by averaging the 4 images (Fig.~\ref{fig:polarization_included}, (b)). To showcase the results of the depth map trained with different depth cameras, we include warped depth images from each depth camera into the polarization camera coordinates using the extrinsic between the two cameras and its depth image (Fig.~\ref{fig:polarization_included}, (d-g)). These can be additionally used for RGBD-based depth completion research. On top of that, we include extra information, such as instance map (Fig.~\ref{fig:polarization_included}, (c)) to help train or validate pipelines for categorical level tasks, accurate 6d pose of the camera as the 4x4 matrix obtained from the robotic arm, extrinsic transformation between cameras as 4x4 matrices and camera intrinsics as 3x3 matrix.

\subsection{D-ToF Camera}
\label{subsec:dtof_camera}

\begin{figure*}[!htbp]
 \centering
    \includegraphics[width=\linewidth]{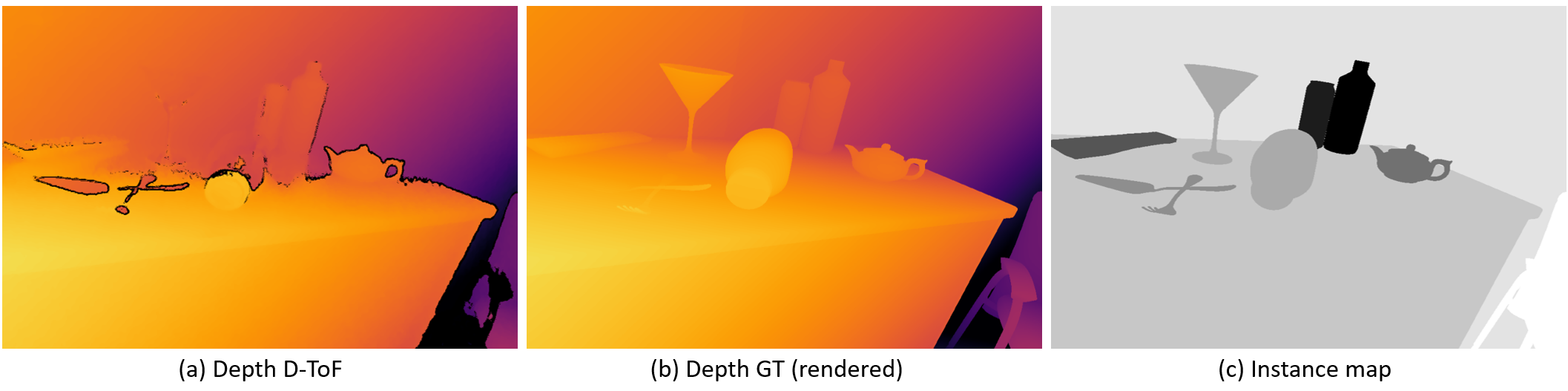}
    \caption{Example of the images included for the D-ToF camera: its depth map (left), ground truth depth (centre) and an object instance label map (right).}
    \label{fig:dtof_included}
\end{figure*}

Fig.~\ref{fig:dtof_included} shows an examples of images included for the D-ToF camera. Direct ToF (D-ToF) camera senses the depth information of its surrounding by emitting the infrared signal and measures the difference in time between emitted and received signal. The quality of this modality highly depends on the reflection of the signal. It often suffers from specific physical noise such as Multi-Path-Interference (MPI) or strong material dependent artefacts (Fig.~\ref{fig:dtof_mpi}). For D-ToF camera, we provide the depth map from the camera (Fig.~\ref{fig:dtof_included}, (a)) as well as its rendered groud truth depth map (Fig.~\ref{fig:dtof_included}, (b))  such that one can also research on D-ToF refinement pipelines to reduce such errors. As in the polarization camera, we include extra information such as instance label map (Fig.~\ref{fig:dtof_included}, (c)), camera pose, intrinsic and extrinsics of the camera as well.

\subsection{I-ToF Camera}
\label{subsec:itof_camera}

\begin{figure*}[!htbp]
 \centering
    \includegraphics[width=\linewidth]{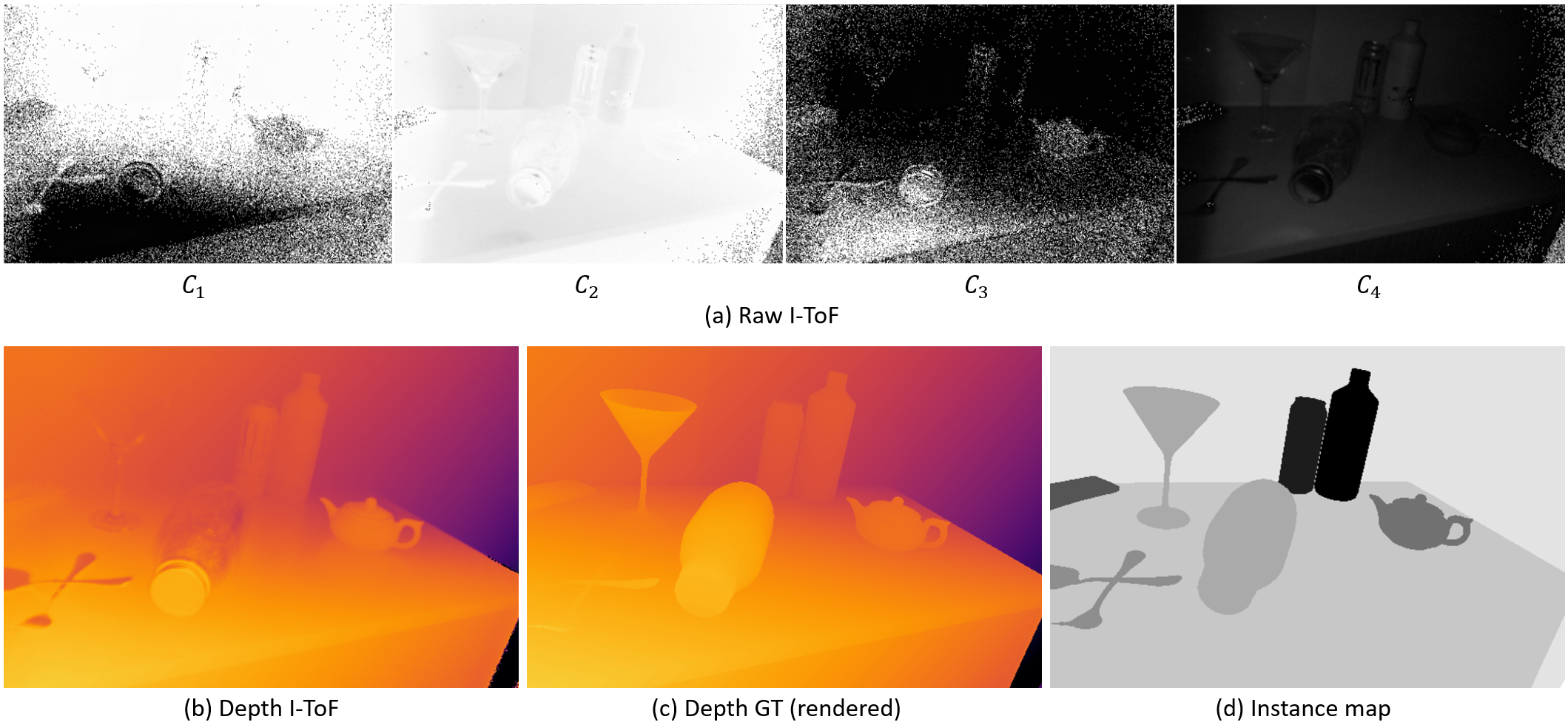}
    \caption{Example of the images included for the I-ToF camera.}
    \label{fig:itof_included}
\end{figure*}

Fig.~\ref{fig:itof_included} shows image examples for the I-ToF camera. Indirect ToF (I-ToF) cameras sense the depth information of their surrounding by emitting a frequency modulated signal and measuring the return signal. Unlike Direct ToF (D-ToF), I-ToF cameras do not calculate the time difference to infer the depth. Instead, the camera correlates the returning signal with phase-shifted emitting signals to generate 4 different measurements, called correlation images. These are measured as sinus functions of distance ($\left( \sin(d),\cos(d),-\sin(d),-\cos(d) \right) = \left( c_{1}, c_{2}, c_{3}, c_{4} \right)$ in Fig.~\ref{fig:itof_included}, (a)). Either arc-tangent formula or convolutional neural network can be used to extract depth information from the correlation images. As I-ToF modality also relies on the reflection of the signal like in D-ToF, it suffers from the similar artefacts, such as MPI and material dependent artefacts (compare qualitative results of the test scenes in Figs.~\ref{fig:qual_scene12_1}, ~\ref{fig:qual_scene13_1} and ~\ref{fig:qual_scene14_1}). Here, we provide raw correlation images and depth map from the camera (see Fig.~\ref{fig:itof_included}, (a,b)) as well as its rendered ground truth depth (Fig.~\ref{fig:itof_included}, (c)) such that one can train I-ToF depth improvement pipelines either from raw signal or from I-ToF depth itself. As the other cameras, extras such as instance map (Fig.~\ref{fig:itof_included}, (d)), camera pose, intrinsic and extrinsics are included.

\subsection{Active Stereo Camera}
\label{subsec:active_stereo_camera}

\begin{figure*}[!htbp]
 \centering
    \includegraphics[width=\linewidth]{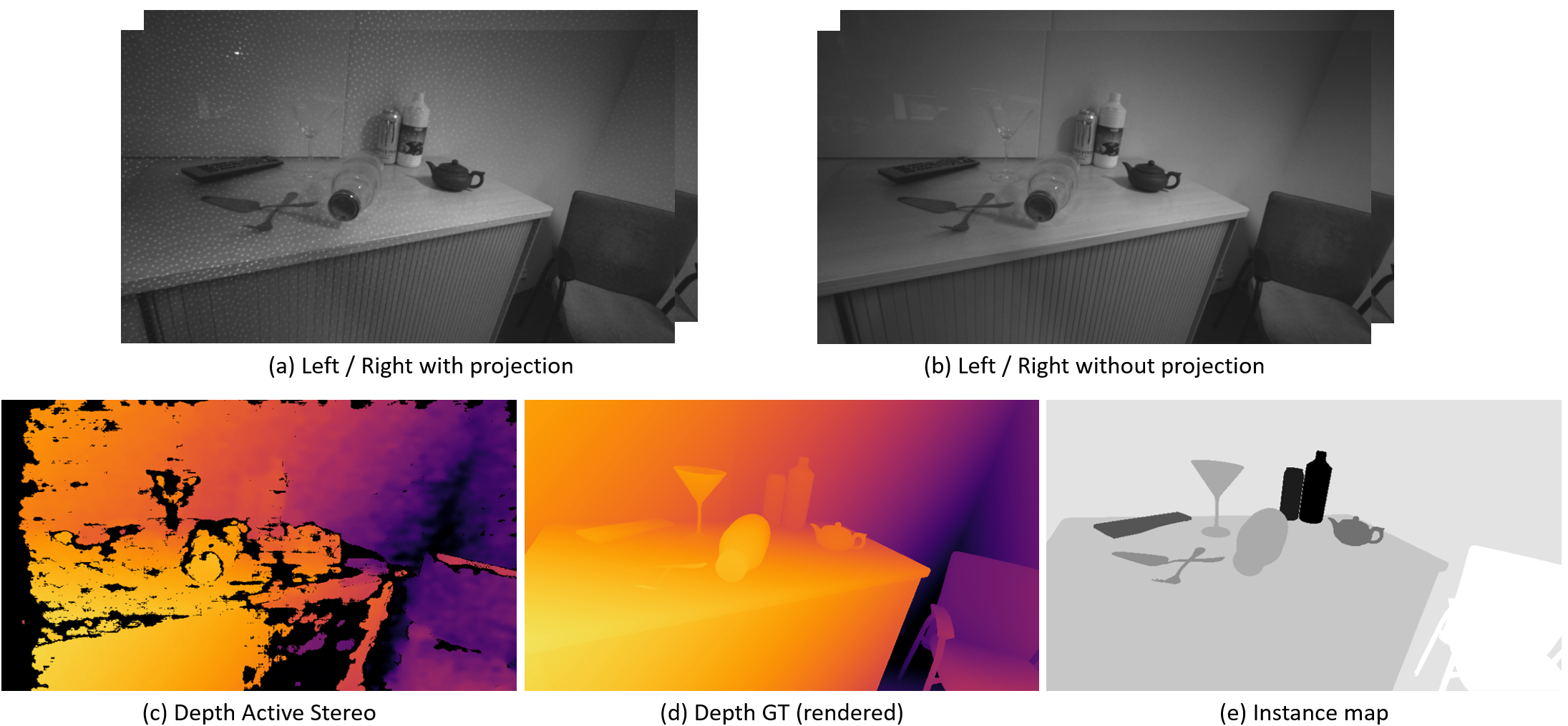}
    \caption{Example of the images included for the Active Stereo camera.}
    \label{fig:d435_included}
\end{figure*}

Fig.~\ref{fig:d435_included} shows the examples of images included for the Active Stereo camera. Stereo depth estimation infers depth using geometrical constraints from epipolar geometry and triangulates the depth map from the disparity between left and right cameras. As the disparity is calculated via matching on the image itself, the stereo based depth estimation methods suffers less on the specific material, but they suffer from other aspects such as stereo occlusion and large texture-less regions. Active projection (Active Stereo) is used to overcome this issue. We provide both active and passive stereo left / right images (Fig.~\ref{fig:d435_included}, (a),(b)) and raw depth from the camera (active,  Fig.~\ref{fig:d435_included}, (c)) as well as the rendered ground truth (Fig.~\ref{fig:d435_included}, (d)). This allows to use our dataset to improve stereo methods from either passive or active stereo and also depth refinement pipelines. Similar to the other cameras, extras such as instance map (Fig.~\ref{fig:d435_included}, (e)), camera pose, intrinsic and extrinsics are included.

\section{Error Analysis on Different Modality}
\label{sec:Error_analysis}
In this section, we show specific error on each depth modality to show the implication of the depth quality when the given modality is used as the ground truth, as well as advantage of using our rendered depth as the ground truth.

\subsection{D-ToF Camera}
\label{subsec:dtof_camera_error}
\begin{figure*}[!htbp]
 \centering
    \includegraphics[width=\linewidth]{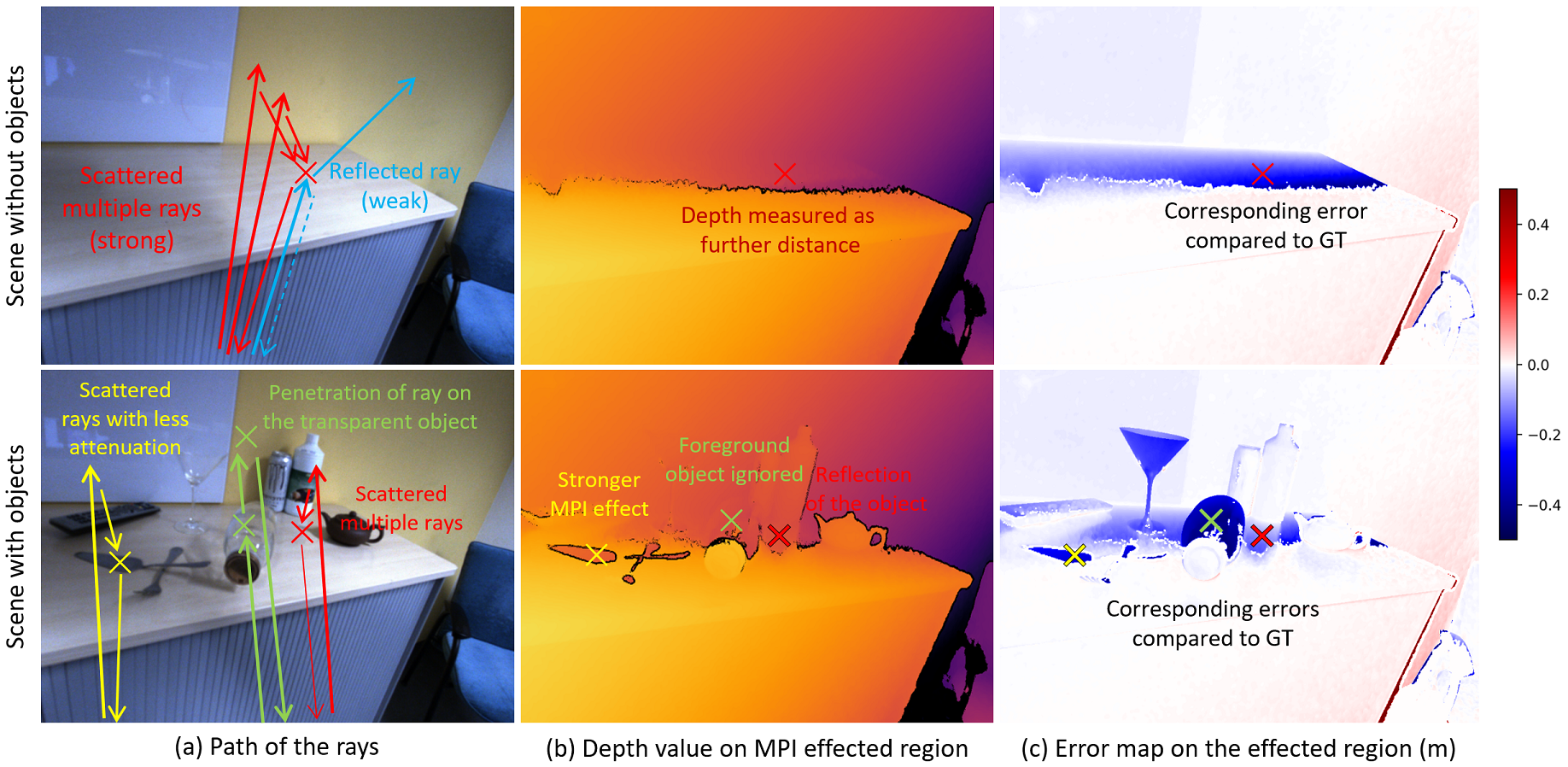}
    \caption{Detailed ray paths with MPI and surface material induced error on D-ToF modality. While D-ToF produces dense and sharp depth, its quality is highly dependent on the surface material and the incident angle.}
    \label{fig:dtof_mpi}
\end{figure*}

\begin{figure*}[!htbp]
 \centering
    \includegraphics[width=\linewidth]{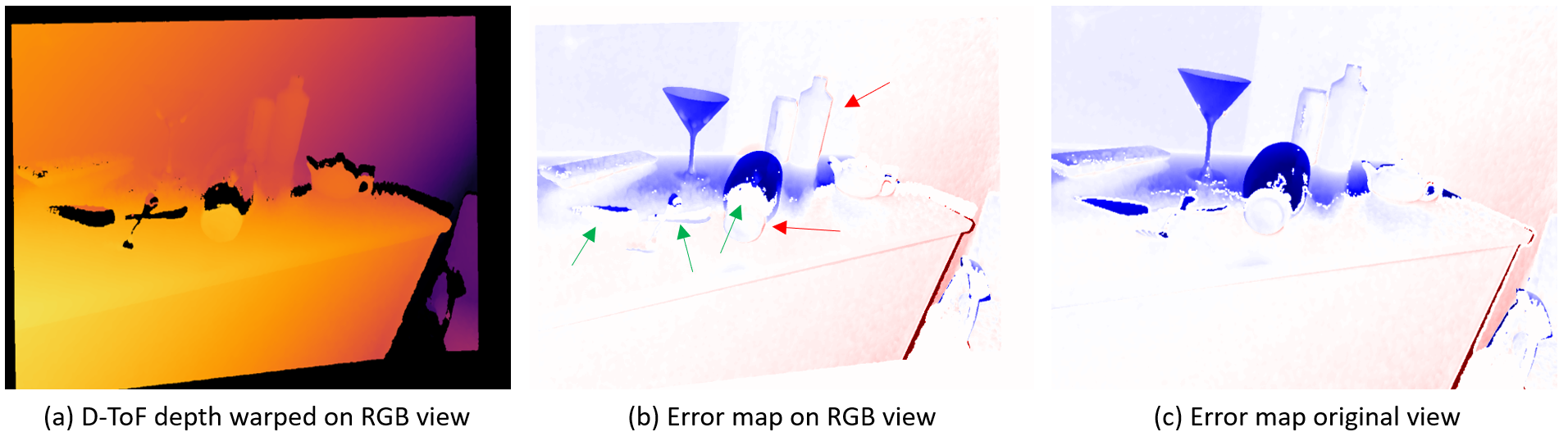}
    \caption{Error after warping D-ToF into RGB view. Slight errors are introduced on some edges (red) while expansion of the invalid area helps to invalidate on the reflective objects (green).}
    \label{fig:dtof_aligned}
\end{figure*}

As mentioned in Subsec.~\ref{subsec:dtof_camera}, D-ToF modality suffers from its own reflection-based nature, such as MPI and material dependent artefact. When the angle of the surface normal of the scene is close to the incident angle of the infrared signal, the strength of the reflected signal becomes weak due to scattering effects (Fig.~\ref{fig:dtof_mpi}, (a) blue arrow) while multiple scattered signals from the other surfaces which has more traveling distance are received and with stronger strength (Fig.~\ref{fig:dtof_mpi}, (a) red arrow) and interfere with the original signal (MPI), producing wrong measurement of the depth on the area with further distance which looks like reflection or shadow of the object to the surface (Fig.~\ref{fig:dtof_mpi}, (b) red marker). This effect can be intensified when the surface material is reflective, which gives even stronger artefact as its reflective surface bounces even weaker and noisier signal with less attenuation (Fig.~\ref{fig:dtof_mpi}, (a,b) yellow arrow\&marker). On the other hands, when the surface material is transparent, the emitted infrared signal rather goes through the object in the both ways  (Fig.~\ref{fig:dtof_mpi}, (a) green arrow) which at the end ignores the object and the sensor produce the depth value as similar level as its background (Fig.~\ref{fig:dtof_mpi}, (b) green marker - material dependent artefact). Quality of the depth map degrades slightly around some boundaries after warping into the RGB frame (Fig.~\ref{fig:dtof_aligned}, (b), red), while the invalid regions actually helps to invalidate more area on wrong depth especially on the reflective objects  (Fig.~\ref{fig:dtof_aligned}, (b), green) , which might become beneficial when it is used in the training.

\subsection{I-ToF Camera}
\label{subsec:itof_camera_error}
\begin{figure*}[!htbp]
 \centering
    \includegraphics[width=\linewidth]{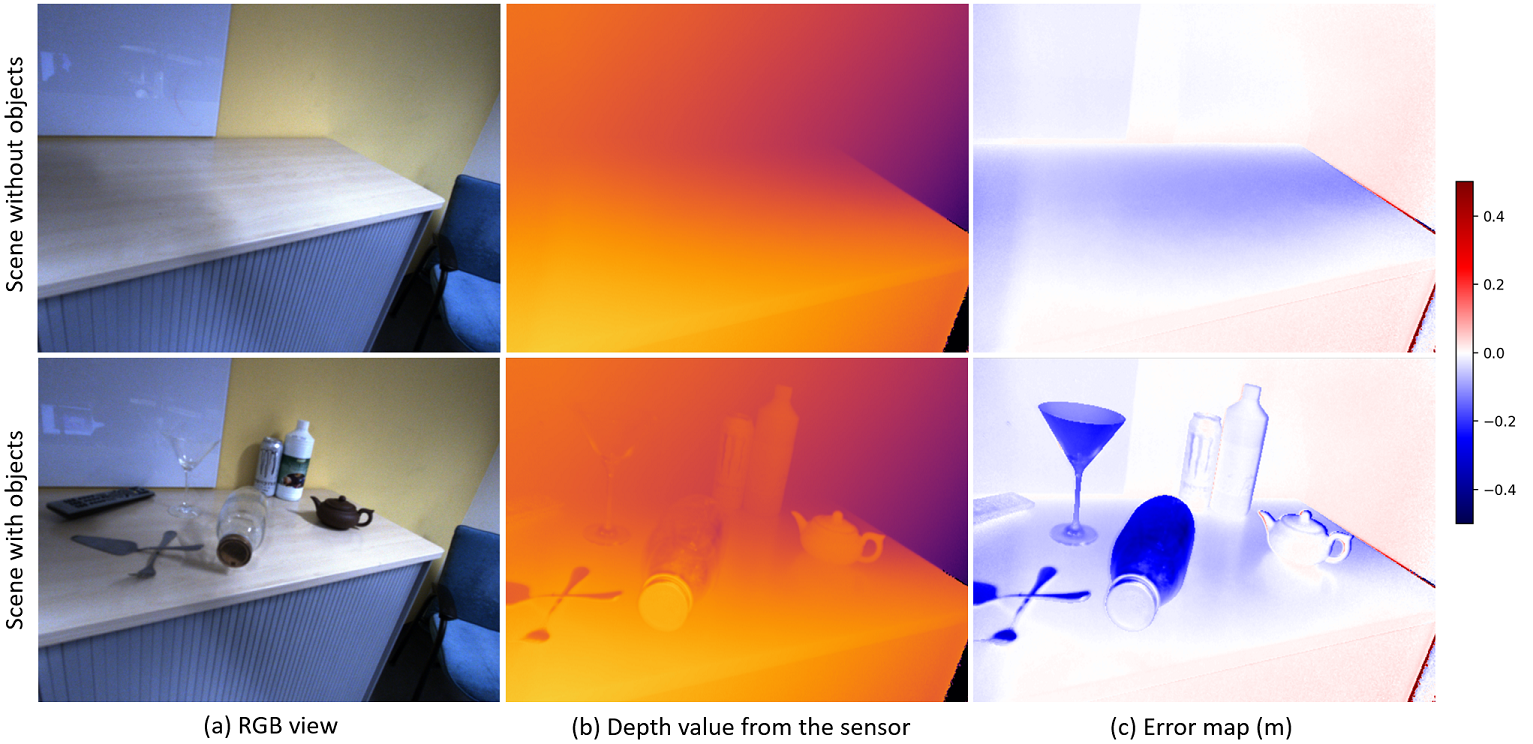}
    \caption{Depth quality from I-ToF camera. I-ToF modality suffers from same type of artefect as D-ToF. While depth map itself is more sense and suffers less from MPI artefact on the table.}
    \label{fig:itof_mpi}
\end{figure*}

\begin{figure*}[!htbp]
 \centering
    \includegraphics[width=\linewidth]{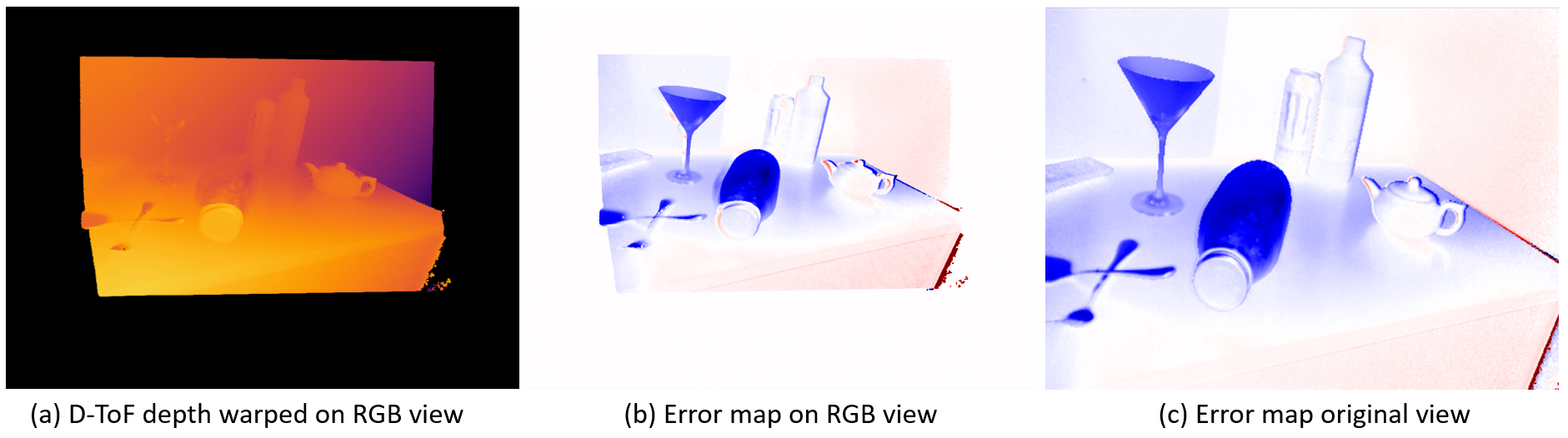}
    \caption{Error after warping I-ToF into RGB view. Not like D-ToF, most of depth error exists without being invalidated, which might introduce more error when it used as GT during the training.}
    \label{fig:itof_aligned}
\end{figure*}

As mentioned in Subsec.~\ref{subsec:itof_camera}, I-ToF modality suffers by its own reflection based nature as well similar to D-ToF, such as MPI and material dependent artefact (Fig.~\ref{fig:itof_mpi}. Although the quality of depth itself seems better as the depth itself is more dense (with less invalid region) and amount of the artefacts are less, it is hard to say I-ToF modality is better than D-ToF as these two camera are in different price range and power level. Also less invalid area but rather with wrong depth didn't help invalidating depth (Fig.~\ref{fig:itof_aligned}) not like in D-ToF case, which could result in artefact in the prediction when it is used as GT during the training.

\subsection{Active Stereo Camera}
\label{subsec:active_stereo_camera_error}
\begin{figure*}[!htbp]
 \centering
    \includegraphics[width=\linewidth]{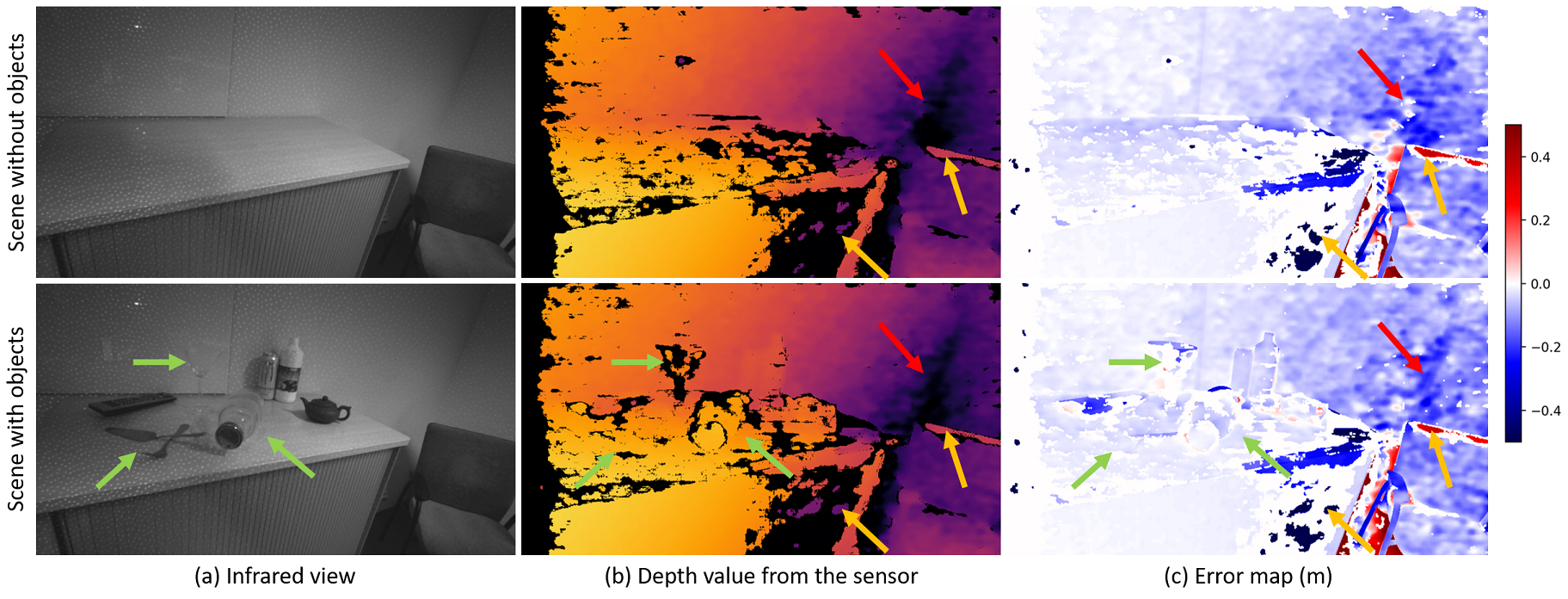}
    \caption{Depth quality from Active Stereo camera. While depth map suffers less on the challenging material, quality of depth itself is far behind either of ToF modality in multiple aspects, such as sharpness, variance, sparsity.}
    \label{fig:d435_not_aligned}
\end{figure*}

\begin{figure*}[!htbp]
 \centering
    \includegraphics[width=\linewidth]{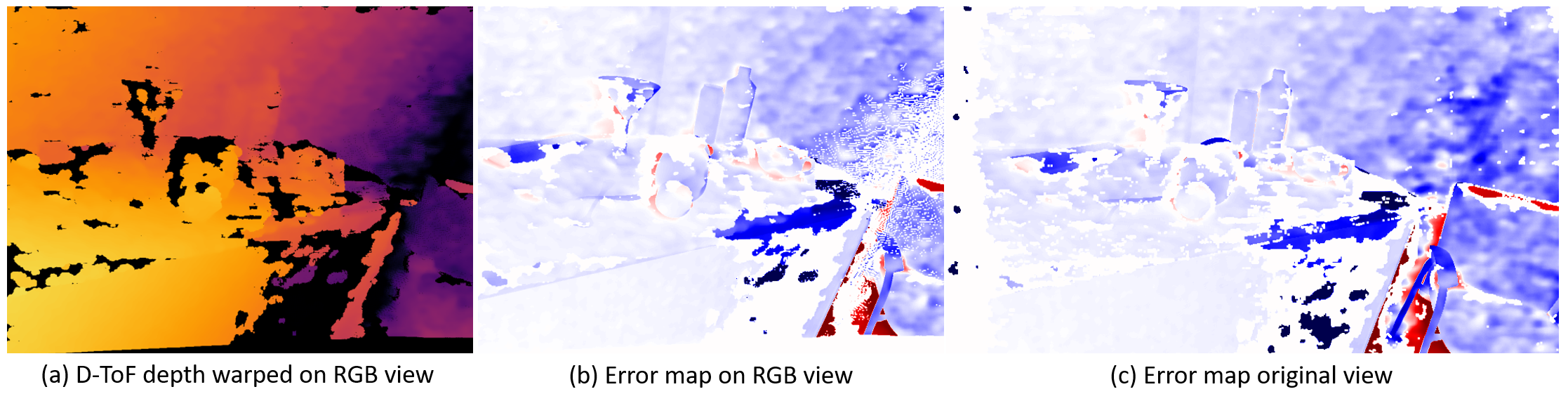}
    \caption{Error after warping Active Stereo into RGB view. Note that there isn't significant change in the depth quality after the warping.}
    \label{fig:d435_aligned}
\end{figure*}

As the stereo camera uses left and right matching with photoelectric cue, depth map suffers less on the challenging material as the projection can be visible on the surface as well as left-right check can be performed to invalidate region with the wrong depth. For this reason, depth on glass or the reflective object is significantly more accurate compared to either of ToF modality (Fig.~\ref{fig:d435_not_aligned}, green arrow). On the other hands, due to its nature of pattern projection far distance that depth quality gets worsen as the scene gets further (Fig.~\ref{fig:d435_not_aligned}, red arrow) the projection pattern gets attenuated and spread in the far distance. Moreover, the depth map in general is more blurry, jittery, sparse and has wrong values on some regions without being invalidated (Fig.~\ref{fig:d435_not_aligned}, orange arrow) which can introduce negative influence when it is used as GT, such as blurriness and depth jittering. Error introduced by warping is trivial (Fig.~\ref{fig:d435_aligned}) as the original depth map is already blurry and sparse.

\section{Qualitative and detailed quantitative evaluation of supervised training}
\label{sec:extra_evaluation}
\subsection{Qualitative predictions}
\begin{figure*}[!ht]
 \centering
    \includegraphics[width=\linewidth]{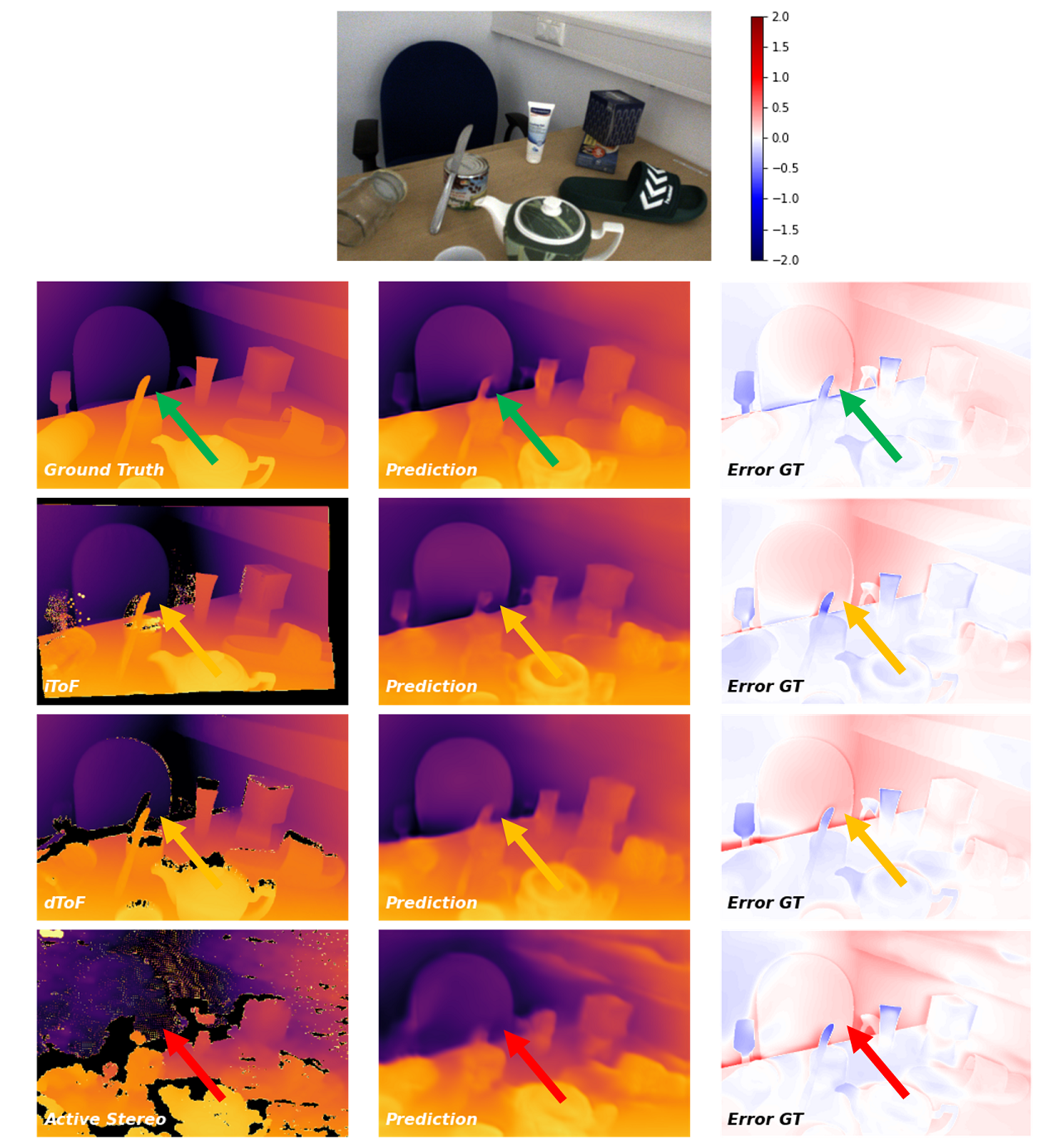}
    \caption{Qualitative evaluation on test scene 1. Each depth modality, the network prediction when trained with supervision of each modality and the error are shown as qualitative evaluation.}
    \label{fig:qual_scene12_1}
\end{figure*}

\begin{figure*}[!ht]
 \centering
    \includegraphics[width=\linewidth]{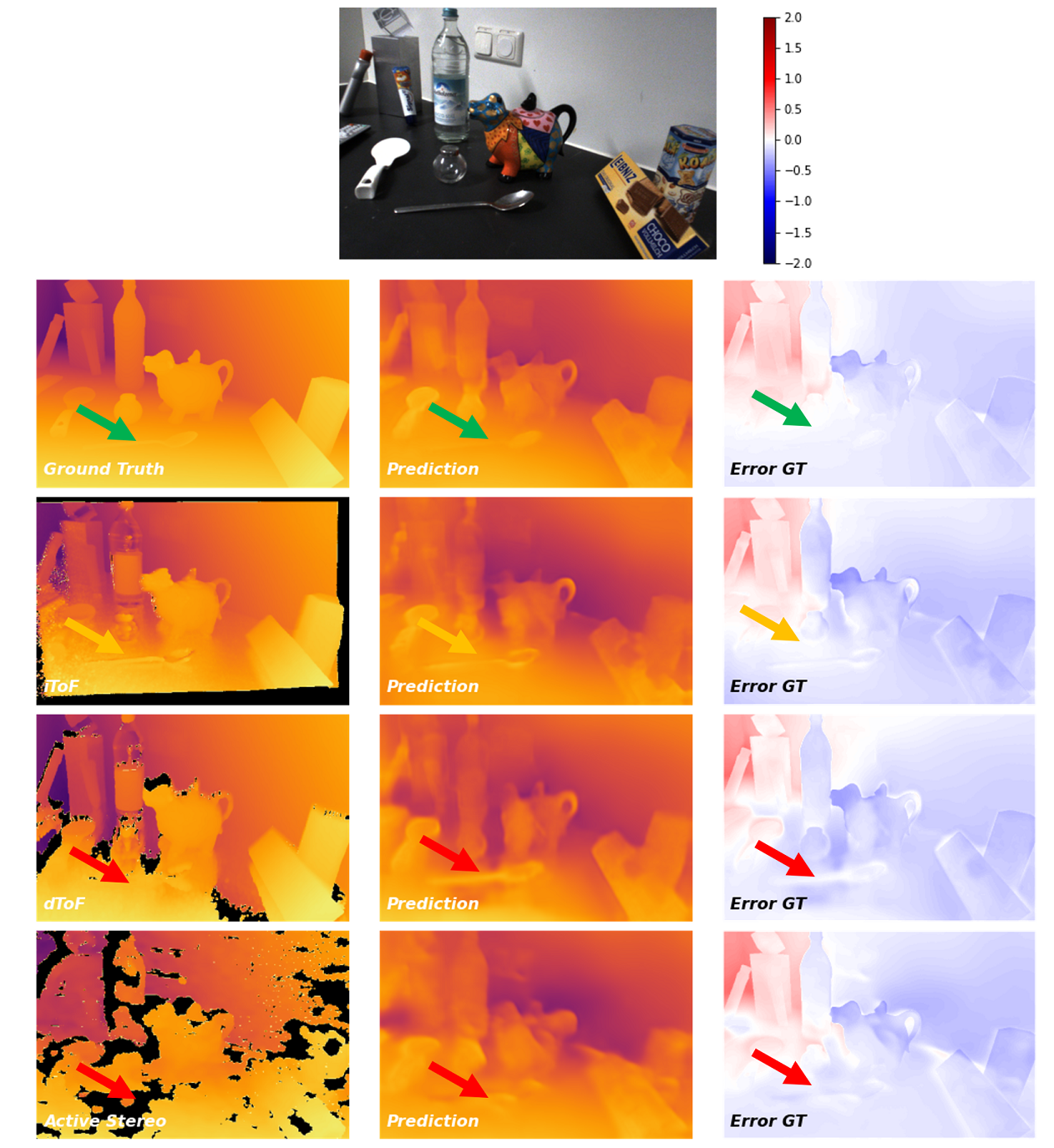}
    \caption{Qualitative evaluation on test scene 2. Each depth modality, the network prediction when trained with supervision of each modality and the error are shown as qualitative evaluation.}
    \label{fig:qual_scene13_1}
\end{figure*}

\begin{figure*}[!ht]
 \centering
    \includegraphics[width=\linewidth]{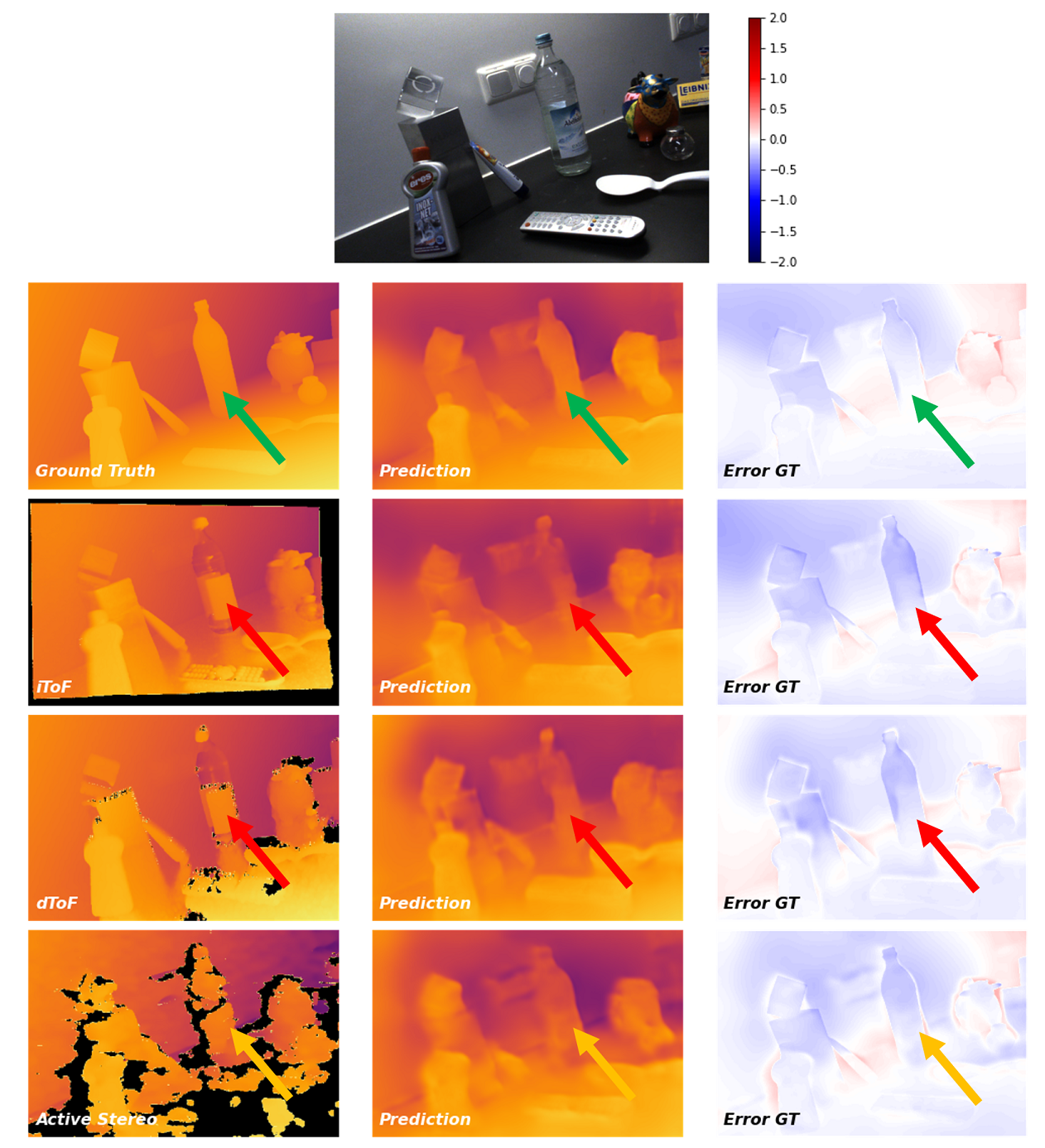}
    \caption{Qualitative evaluation on test scene 3. Each depth modality, the network prediction when trained with supervision of each modality and the error are shown as qualitative evaluation.}
    \label{fig:qual_scene14_1}
\end{figure*}

Figures~\ref{fig:qual_scene12_1},~\ref{fig:qual_scene13_1} and~\ref{fig:qual_scene14_1} show predictions on exemplary frames of the test scenes 1, 2 and 3, together with the different sensor modalities and the error plot of the prediction compared against the ground truth. The training with rendered ground truth generally performs best. Both ToF sensors show incorrect depth values for reflective or transparent objects which also translates to incorrect predictions in these areas (compare Fig.~\ref{fig:qual_scene12_1}. The predictions when training with active stereo as supervision are more blurry and show less distinct edges at depth boundaries when compared to other modalities, which may arise from many depth pixels being invalidated by the sensors around such boundaries (compare Fig.~\ref{fig:qual_scene13_1}). The very challenging test scene 3 with bright lighting and many unseen objects is difficult to predict for all training setups (compare Fig.~\ref{fig:qual_scene14_1}. We can see similar artifacts as described above. Additionally, the unseen trophy object with partly reflective and partly transparent material shows large errors for the sensor inputs as well as for its predictions. The desk surface is also incorrectly captured by the d-ToF sensors due to large reflections and MPI from the background.

\subsection{Quantitative evaluation}
\paragraph{Test scenes. }
Table~\ref{tab:depth_supervision_results_additional} summarizes the extensive quantitative evaluation of the supervised training with different depth modalities as supervision signal for different test scenes. Test scene 1 has a similar background compared to the training scenes and includes additional unseen objects. The scene is also observed from viewing angles that differ significantly from the training data. The background in test scene 2 is only partly observed in the training data and it includes mostly unseen objects. Test scene 3 is similar to test scene 2, but with a modified object layout and difficult lighting in the background from an additional bright light source above the scene. The additional test set with (partly) seen scenes is an additional test split which includes the first 10 frames of each training sequence. Please note that these frames have not been used during training. Here, we first test all predictions against the rendered ground truth and additionally on each individual respective modality to highlight the overfitting issue of invalid ground truth from each modality.
The results suggest that overall the supervision with accurate rendered ground truth achieves to generalize best for (mostly) unknown scenes. It is noticeable, that the active strereo achieves to produce good predictions for transparent objects and also performs well for reflective ones. The i-ToF and d-ToF predictions suffer from incorrect ground truth values for such objects.

\paragraph{Overfitting on (partly) seen scenes. }
The (partly) seen scene shows generally lower overall errors for all modalities as compared to the (mostly) unseen test scenes 1,2, and 3. Again, the active stereo can provide decent depth supervision for reflective and transparent objects, where the ToF sensors cannot provide valid depth. The prediction of the background of the scene performs worst for the active stereo, as the textureless wall is still problematic for the sensor.

When testing on the respective modality itself, the overfitting issue due to incorrect depth values of the sensor becomes apparent. It can be noticed, that for objects where the respective sensor cannot yield accurate depth values (e.g. transparent objects for i-ToF or reflective objects for d-ToF), the errors are significantly lower, indicating overfitting to the specific sensor modality.

The training and testing scripts together with the definition of each test scene split will be provided with the camera-ready version of the paper.

\begin{table*}[!ht]
\centering
\caption{\textbf{Depth prediction comparison when training with different modalities and tested on different unseen scenes and seen scenes. }(Top) Evaluation against GT of depth predictions on the test set with dense supervision from different depth modalities. (Bottom) Predictions evaluated on respective modality. 
Error is first computed in meters scale and here reported as Sq.Rel. and RMSE in mm.
}
\begin{adjustbox}{angle=90}
\footnotesize
\resizebox{1.2\textwidth}{!}{
\begin{tabular}{ll|cc|cccc|cccccc}
\shline
& Mask&\multicolumn{2}{c|}{Full Scene} & \multicolumn{2}{c}{Background} & \multicolumn{2}{c|}{All Objects}  & \multicolumn{2}{c}{Textured} & \multicolumn{2}{c}{Reflective} & \multicolumn{2}{c}{Transparent} \\
\hline
&Metric&Sq.Rel. & RMSE&Sq.Rel. & RMSE&Sq.Rel. & RMSE&Sq.Rel. & RMSE&Sq.Rel. & RMSE&Sq.Rel. & RMSE \\
\shline
\multirow{4}{*}{\rotatebox[origin=c]{-90}{Test 1}}     & iToF & 
24.78  &  148.09
 & 
22.25  &  151.07
 & 
29.62  &  123.19
 & 
16.47  &  99.08
 & 
102.79  &  214.60
 & 
44.29  &  134.44
\\
                                & dToF & 
24.23  &  151.72
 & 
23.74  &  159.28
 & 
22.85  &  110.88
 & 
16.22  &  101.12
 & 
57.14  &  148.61
 & 
30.23  &  107.23
\\
                                & Active Stereo & 
32.15  &  173.72
 & 
33.84  &  184.16
 & 
22.23  &  116.57
 & 
19.55 &  114.07
 & 
64.27  &  167.71
 & 
12.92  &  69.49
  
 \\
                                & GT & 
22.16  &  143.45
 & 
22.27  &  150.77
 & 
19.09 &  103.77
& 
15.27  &  96.51
 & 
61.96  &  161.42
 & 
12.28  &  68.74
\\
\shline\\
\shline

\multirow{4}{*}{\rotatebox[origin=c]{-90}{Test 2}}     & iToF & 
27.42  &  123.79
 & 
22.66  &  116.86
 & 
39.85  &  139.67
 & 
48.66  &  144.92
 & 
16.15  &  99.44
 & 
25.15  &  122.25
\\
                                & dToF & 
23.00  &  115.40
 & 
21.18  &  113.27
 & 
27.89  &  119.59
 & 
30.00  &  112.92
 & 
15.81  &  90.89
 & 
23.73  &  117.72
\\
                                & Active Stereo & 
25.94  &  124.17
 & 
25.50  &  126.28
 & 
27.18  &  117.04
 & 
32.81  &  121.24
 & 
16.40  &  101.86
 & 
15.73  &  95.27
\\
                                & GT & 
19.96  &  108.91
 & 
19.25  &  109.94
 & 
21.75  &  104.94
 & 
25.85  &  106.22
 & 
17.05  &  104.96
 & 
9.62  &  73.97
\\
                                
\shline\\
\shline

\multirow{4}{*}{\rotatebox[origin=c]{-90}{Test 3}}     & iToF & 
36.82  &  152.51
 & 
35.92  &  153.26
 & 
38.75  &  147.14
 & 
34.09  &  127.51
 & 
20.21  &  110.85
 & 
55.09  &  183.14
\\
                                & dToF & 

32.99  &  145.50
 & 
35.64  &  153.07
 & 
25.90  &  120.35
 & 
19.92  &  96.01
 & 
21.59  &  105.41
 & 
37.26  &  149.66
\\
                                & Active Stereo & 
31.63  &  141.77
 & 
35.24  &  151.37
 & 
22.44  &  110.42
 & 
23.47  &  106.63
 & 
14.49  &  94.51
 & 
21.21  &  109.53
\\
                                & GT & 
27.56  &  131.77
 & 
28.95  &  137.73
 & 
23.62  &  112.23
 & 
25.95  &  108.67
 & 
18.94  &  108.45
 & 
14.99  &  90.66
\\
                                
\shline\\
\shline

\multirow{4}{*}{\rotatebox[origin=c]{-90}{Test Seen}}     & iToF & 
9.87  &  77.99
 & 
4.62  &  57.10
 & 
33.91  &  133.46
 & 
6.18  &  60.48
 & 
35.65  &  119.76
 & 
91.30  &  224.27
\\
                                & dToF & 

15.43  &  93.31
 & 
11.62  &  79.89
 & 
31.12  &  123.97
 & 
4.40  &  51.91
 & 
17.42  &  82.29
 & 
89.19  &  212.55
\\
                                & Active Stereo & 
9.43  &  88.30
 & 
9.28  &  88.24
 & 
9.11  &  75.21
 & 
6.32  &  65.54
 & 
12.98  &  65.73
 & 
16.62  &  98.75
\\
                                & GT & 
1.12  &  28.81
 & 
0.71  &  24.41
 & 

2.65  &  40.41
&
1.83  &  34.89
 & 
2.16  &  29.55
 & 
5.02  &  52.43
\\
\shline\\
\\
\\
\multicolumn{6}{l}{Tested on Modality:}
\\
\shline
\multirow{4}{*}{\rotatebox[origin=c]{-90}{Test Seen}}     & iToF & 
8.34  &  52.29
 & 
8.57  &  50.00
 & 

7.01  &  58.85
&
3.80  &  43.44
 & 
23.28  &  95.38
 & 
13.69  &  65.41
\\
                                & dToF & 

8.05  &  50.43
 & 
6.82  &  45.50
 & 
 
13.52  &  66.34
&
9.00  &  54.15
 & 
30.91  &  87.71
 & 
27.92  &  87.32
\\
                                & Active Stereo & 
39.25  &  101.76
 & 
40.87  &  102.29
 & 
 
30.32  &  90.00
&
32.24  &  90.49
 & 
23.36  &  72.21
 & 
37.25  &  101.23
\\
                                & GT & 
1.12  &  28.81
 & 
0.71  &  24.41
 & 
 
2.65  &  40.41
&
1.83  &  34.89
 & 
2.16  &  29.55
 & 
5.02  &  52.43
\\
\shline

\end{tabular}
}
\end{adjustbox}
\label{tab:depth_supervision_results_additional}
\end{table*}

\end{document}